\documentclass[11pt]{article}
\usepackage{acl}

\usepackage{times}
\usepackage{latexsym}

\usepackage{makecell}
\usepackage[T1]{fontenc}
\usepackage[utf8]{inputenc}
\usepackage{microtype}
\usepackage{inconsolata}
\usepackage{xcolor}
\usepackage{adjustbox}

\usepackage{graphicx}          
\usepackage{subcaption}        
\usepackage{caption}           
\usepackage{graphicx}
\usepackage{lipsum} 

\usepackage{multirow}
\usepackage{amsmath,amssymb}
\usepackage{wrapfig}  

\usepackage{tikz}
\usetikzlibrary{positioning,shapes.misc,shadows.blur,arrows.meta}

\usepackage{mathtools}
\usepackage{enumitem}
\usepackage{booktabs}
\usepackage{siunitx}

\usepackage{tabularx}
\usepackage{array}

\usepackage{listings}
\lstdefinestyle{jsonstyle}{
  basicstyle=\ttfamily\small,
  columns=fullflexible,
  breaklines=true,
  frame=none,
  showstringspaces=false
}

\usepackage{colortbl} 
\usepackage{xcolor} 
\usepackage[most]{tcolorbox}
\definecolor{backblue}{RGB}{210, 230, 250}
\definecolor{backgreen}{RGB}{226, 240, 217}
\definecolor{backred}{RGB}{255, 223, 223}
\newcommand{\highblue}{\cellcolor{backblue}}
\newcommand{\highgreen}{\cellcolor{backgreen}}

\definecolor{deepgreen}{RGB}{0, 70, 0}

\usepackage{pifont}
\newcommand{\cmark}{\textcolor[rgb]{0.0, 0.6, 0.0}{\ding{51}}} 
\newcommand{\xmark}{\textcolor[rgb]{0.7, 0.0, 0.0}{\ding{55}}} 

\lstdefinestyle{python}{
    language=Python,
    basicstyle=\fontsize{7}{8.5}\ttfamily,
    keywordstyle=\color{blue},
    commentstyle=\color{gray},
    stringstyle=\color{black},
    showstringspaces=false,
    breaklines=true,
    breakindent=0pt,
    breakatwhitespace=false,
    escapeinside={(*@}{@*)},
    morecomment=[s]{"""}{"""},
}

\newcounter{programbox}

\title{ShopSimulator: Evaluating and Exploring RL-Driven LLM Agent for Shopping Assistants}

\author{Pei Wang\thanks{First three authors contributed equally. $^{\dag}$ Corresponding Author: Yanan Wu.}~, Yanan Wu$^{* \dag}$, Xiaoshuai Song$^{*}$, Weixun Wang, Gengru Chen \\
\bf{Zhongwen Li}, Kezhong Yan, Ken Deng, Qi Liu, Shuaibing Zhao, Shaopan Xiong\\
\bf{Xuepeng Liu}, Xuefeng Chen, Wanxi Deng, Wenbo Su, Bo Zheng \hfill \\
  Alibaba Group \hfill
  \\ \texttt{\{yupei.wp, lixing.wyn, songxiaoshuai.sxs\}@alibaba-inc.com}}

\begin{document}
\maketitle
\begin{abstract}
Large language model (LLM)‑based agents are increasingly deployed in e‑commerce shopping. 
To perform thorough, user‑tailored product searches, agents should interpret personal preferences, engage in multi‑turn dialogues, and ultimately retrieve and discriminate among highly similar products.
However, existing research has yet to provide a unified simulation environment that consistently captures all of these aspects, and always focuses solely on evaluation benchmarks without training support.
In this paper, we introduce ShopSimulator, a large‑scale and challenging Chinese shopping environment. 
Leveraging ShopSimulator, we evaluate LLMs across diverse scenarios, finding that even the best‑performing models achieve less than 40\% full‑success rate. Error analysis reveals that agents struggle with deep search and product selection in long trajectories, fail to balance the use of personalization cues, and to effectively engage with users.
Further training exploration provides practical guidance for overcoming these weaknesses, with the combination of supervised fine‑tuning (SFT) and reinforcement learning (RL) yielding significant performance improvements.  Code and data are released at \url{https://github.com/ShopAgent-Team/ShopSimulator}.
\end{abstract}

\section{Introduction}
In modern e‑commerce platforms, personalized and interactive product search have become key components of the user experience. Beyond matching products to explicit queries, real‑world scenarios often involve users who provide vague, partial, or evolving goals. 
An effective shopping assistant should communicate with the user to clarify these goals, infer preferences from user profiles and historical behavior, then search and distinguish similar products to offer the most suitable options.
Recent advances in large language models (LLMs) allow LLMs to integrate interaction, reasoning, and search into a unified workflow, making LLM‑based agents promising candidates for the next generation of shopping assistants in e‑commerce~\citep{yao2022webshop, wu2024survey, xu2024survey}.

\begin{table*}[t!]
    \centering
    \small
\setlength{\tabcolsep}{1mm}{
\begin{adjustbox}{width=\textwidth, keepaspectratio}
\begin{tabular}{l c c c c c c c c c}
\toprule
\textbf{Environment} & \textbf{Language} & \textbf{Source} & \textbf{\# Domain} & \textbf{\# Product} & \textbf{\# Task} & \textbf{\makecell[c]{Training-Task}} & \textbf{\makecell[c]{Single-Turn}} & \textbf{\makecell[c]{Multi-Turn}}  & \textbf{\makecell[c]{Personalization}} \\
\midrule
WebShop~\citep{yao2022webshop} & English & Amazon & 5 & 1.18M & 12,087 & \cmark & \cmark & \xmark & \xmark \\
DeepShop~\citep{lyu2025deepshopbenchmarkdeepresearch} & English & Amazon & 5 & -- & 600 & \xmark & \cmark & \xmark & \xmark \\
ChatShop~\citep{chen2024chatshopinteractiveinformationseeking} & English & Amazon & 5 & 1.18M & 1,500 & \xmark & \cmark & \cmark & \xmark \\
WebMall~\citep{peeters2025webmallmultishopbenchmark} & English & WooCommerce & 3 & 4.4K & 91 & \xmark & \cmark & \xmark & \xmark \\
ShoppingBench~\citep{wang2025shoppingbenchrealworldintentgroundedshopping} & English & Lazada & 21 & 2.5M & 3,310& \cmark & \cmark & \xmark & \xmark \\
\textbf{ShopSimulator (Ours)} & Chinese & Taobao & 12 & 1.34M & 28,147 & \cmark & \cmark & \cmark & \cmark \\
\bottomrule
\end{tabular}
\end{adjustbox}
}
    \caption{Comparison of ShopSimulator with existing work. 
    Columns indicate: 
    Training Task– whether a training task set is included to support model training; 
    Multi-Turn – whether multi-turn dialogues with users are supported; 
    Personalization – whether user-specific personalization such as user information and historical behavior is supported.
    Note: DeepShop crawls live e-commerce sites in real time, with a variable product count.}
 \label{tab:existing_work}
\end{table*}

To develop such agents, the community has shown growing interest in applying reinforcement learning (RL) techniques to the training of LLMs. Unlike supervised fine‑tuning (SFT) or offline preference optimization (e.g., DPO~\citep{rafailov2023direct}), RL algorithms (e.g., PPO, GRPO~\citep{shao2024deepseekmath}) depend on autonomous exploration within interactive environments, using feedback signals to iteratively refine the agent’s policy.
Although several studies have made progress toward environment‑driven e‑commerce exploration, such as WebShop and DeepShop~\citep{yao2022webshop, lyu2025deepshopbenchmarkdeepresearch}.
However, as shown in Table \ref{tab:existing_work}, they face two key limitations. First, they lack a unified environment that simultaneously captures the personalization, multi‑turn interaction\footnote{In this paper, ``multi-turn'' refers to agent–user interactions, while ``multi-step'' refers to agent–env actions.}, and fine‑grained product discrimination required in realistic e‑commerce settings. Second, they focus primarily on evaluation and provide little or no support for training resources.
These limitations undermine both the reliable assessment of agent performance in realistic user interactions and the effectiveness of these environments for RL‑driven strategy exploration and training\footnote{We introduce the related work in detail in Section \ref{sec:related_work}.}.
To address this gap, we focus on two research questions:

\emph{\textbf{RQ1}: How do current LLM agents perform, and can they serve as reliable shopping assistants in environments characterized by personalized user preferences and multi‑turn interactions?}

\emph{\textbf{RQ2}: In such environments, which training strategies, notably RL compared to SFT, can effectively drive LLMs toward reliable shopping agents?}

As the foundation for exploration, we introduce ShopSimulator, a Chinese e‑commerce sandbox environment grounded in real‑world products and realistic user characteristics, designed to support both the evaluation and training of agents in lifelike shopping scenarios.
In this environment, an agent engages in multi‑turn interactions with the user to clarify purchase intentions, search and browse a sandboxed product catalog, reason over product attributes and user preferences, and ultimately recommend the most suitable item.
To reflect real‑world platforms, ShopSimulator collects over 1.3 million products from Taobao\footnote{\url{www.taobao.com}. We collect data under authorization, as stated in Appendix~\ref{app:ethic_state}.} across 12 domains, with each sub‑category containing highly similar items. 
This requires agents to first perform coarse‑grained retrieval to get candidate groups and then fine‑grained discrimination among similar products.
For user modeling, ShopSimulator uses LLM‑driven role‑playing shoppers equipped with detailed personal profiles including long‑term preferences, demographic attributes, and historical purchase patterns.
To support training, ShopSimulator offers 25K training tasks along with 2.8K evaluation tasks,  covering single‑ and multi‑turn interactions in both personalized and non‑personalized settings. Each task is accompanied by reward signals derived from multiple dimensions,  including categories, attributes, options, and prices.

In response to RQ1, we evaluate a range of advanced LLMs on ShopSimulator test set.
The results reveal: 
(1) Current LLMs are still far from being reliable agents for shopping assistants. Even GPT‑5 achieves 32\% full-success rate.
Although recommendations often meet category and price constraints, they fail to satisfy fine‑grained attribute and option requirements.
(2) From a behavioral perspective, agents often perform redundant retrieval or miss key attributes in retrieval, under‑utilize available results or enforce constraints weakly in product viewing, and make hasty purchase decisions with insufficient user communication in recommendation.
(3) In terms of personalization, failures largely arise from imbalance between underusing and over-reasoning preference information. 

In response to RQ2, we explore RL training for Qwen3‑8B using the ShopSimulator train set. Our main findings are:
(1) The learning paradigms of SFT and RL are complementary, making the ``SFT + RL'' consistently outperform RL alone across all scenarios. SFT injects priors and task workflows into LLMs, while RL further learns preferences and enhances fulfillment of fine-grained needs.
(2) Using the multiplicative strict reward as the RL objective yields consistently better performance than the additive loose reward across all task scenarios, with particularly notable advantages in attribute and option matching. This benefit stems from the bottleneck‑style reward focusing optimization on weaker dimensions.  
(3) Single‑turn tasks see greater improvements after training, whereas multi‑turn tasks remain at around 35\% success rate, highlighting the challenge for LLMs to simultaneously perform personalization, intent clarification, and environment interaction over long trajectories.

In summary, our contributions are threefold:
(1) We introduce ShopSimulator, a sandbox for evaluating and training RL‑driven agents in realistic shopping scenarios, featuring a large product catalog, personalized multi‑turn user modeling, and fine‑grained product differentiation.
(2) We comprehensively evaluate advanced LLMs, revealing a substantial gap to becoming reliable shopping assistants. Detailed error analysis further offers insights into underlying deficiencies and improvement directions.
(3) We explore training strategies on Qwen3‑8B, showing the complementarity of SFT and RL and the advantages of strict rewards, while the varying gains across scenarios
highlight the challenges in multi-turn personalization.

\begin{figure*}[t]
    \centering
    \resizebox{\textwidth}{!}{
    \includegraphics{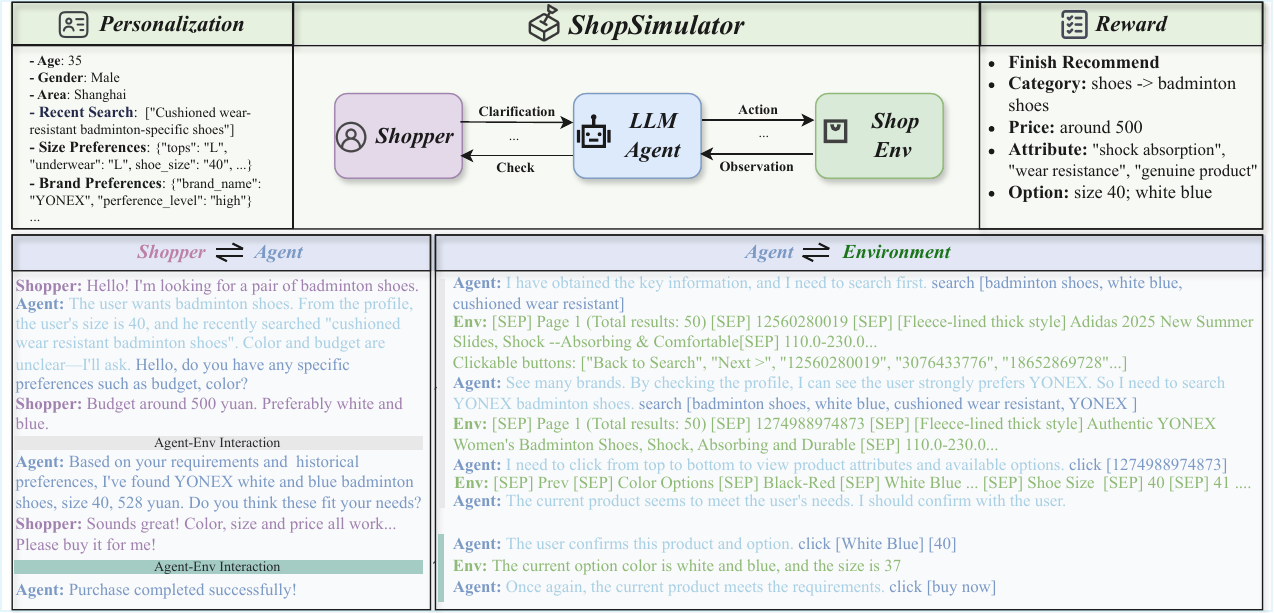}
    }
\caption{Illustration of ShopSimulator. On the one hand, the assistant agent needs to interpret preferences from user profile and communicate with users to understand their shopping needs; on the other hand, the agent iteratively search the database, click and review products, and ultimately to  recommend suitable items. }
    \label{fig:env_main_intro}
\end{figure*}

\section{ShopSimulator}
\label{sec:environment}

\subsection{Task Formulation}
\label{subsec:formulation}

In WebShop and similar environments~\citep{yao2022webshop, lyu2025deepshopbenchmarkdeepresearch}, the shopping task is modeled as a sequential decision process in which an agent interacts with a text-based shopping interface to fulfill a fixed and explicit user goal \(G\). At each timestep \(t\), the policy \(\pi_{\theta}\) generates an action \(a_t\) based on the current observation \(o_t\) and the goal \(G\):  
\begin{equation}
a_t = \pi_{\theta}(o_t, G)\quad o_{t+1} = \mathcal{E}(o_t, a_t),
\end{equation}
where \(\mathcal{E}\) is the environment transition function returning the next observation \(o_{t+1}\). The action space includes issuing textual search queries over the product catalog, browsing search results, viewing product details, and making a purchase. An episode terminates when completing a purchase or reaching a maximum step limit.

\begin{figure*}[t]
    \centering
    \resizebox{\textwidth}{!}{
    \includegraphics{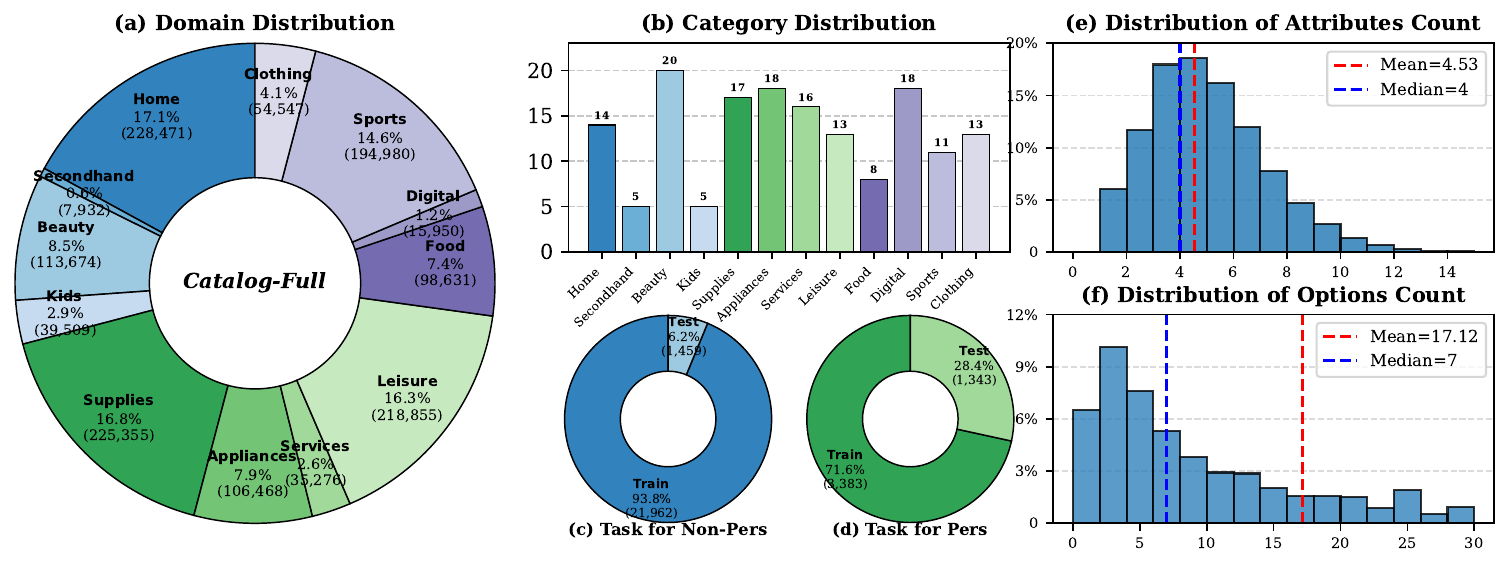}
    }
\caption{Statistics Dashboard of ShopSimulator. 
Fig (a) shows the 12 domains, while Fig(b) shows the number of first-level categories within each domain.
Fig (c) and (d) show the train–test split of instructions for non-personalized and personalized settings. Each instruction can be used in both single-turn and multi-turn dialogues.  
Fig (e), (f) show the average attributes (e.g., unisex) and options (e.g., size–color combinations) per product, respectively. }
    \label{fig:env_info}
\end{figure*}

As shown in Figure \ref{fig:env_main_intro}, ShopSimulator retains the core interaction paradigm of WebShop while introducing three key extensions:
(1) \textbf{Multi‑turn interaction}: 
incorporates an LLM‑simulated user that supports multi‑turn dialogue, requiring the agent to actively clarify underspecified or ambiguous goals;
(2) \textbf{Personalization}: conditions the agent’s policy on structured user profiles and historical behaviors to incorporate user preferences; 
(3) \textbf{Fine‑grained product discrimination}: each smallest sub‑category contains around 120 highly similar items, demanding more precise selection than random sampling.

Formally, the agent’s policy in ShopSimulator now conditions on the current user utterance \(u_t\), and a static user profile \(p\):  
\begin{equation}
a_t = \pi_{\theta}(o_t, u_t, p), \quad
o_{t+1}, u_{t+1} = \mathcal{E}(o_t, a_t)
\end{equation}
where, compared with the single‑turn setting, the multi‑turn mode extends the action space to include direct interaction with the user, enabling the acquisition of additional information or updated needs \(u_{t+1}\); personalization provides the profile \(p\) to support tailored search. To support diverse scenarios, ShopSimulator also maintains single‑turn and multi‑turn non‑personalized settings.

\subsection{Environment Construction}
\label{subsec:env_construct}

\textbf{Overview.}
The construction of ShopSimulator follows a three‑stage pipeline: product catalog collection, task generation, and user modeling.
We first collect the product catalog from a real e‑commerce platform snapshot, applying filtering and sampling to create a high‑quality subset (Catalog‑Fine) and a broad‑coverage dataset (Catalog‑Full). 
Next, each product is paired with a manually annotated instruction, ensuring unambiguous one‑to‑one mapping and thereby instantiating the shopping tasks. 
Finally, to model realistic users, we synthesize structured user profiles, and employ LLM‑simulated shoppers to engage in multi‑turn interactions with the agent over the task\footnote{We provide examples of products, tasks,and personalized user profiles in Appendix~\ref{subapp:example_details}.}.

\textbf{\textit{Product Catalog Collection.}}
We collect real‑world product data from Taobao, the largest Chinese e‑commerce platform.
To ensure data recency, we use a snapshot taken in June 2025.
By computing the exposure frequency of each product, we select the top 50M high‑frequency items as the initial pool.
To achieve a complex and diverse hierarchical coverage, we first discard first‑level categories containing fewer than ten sub‑categories.
For the remaining ones, we employ GPT‑4o to assess both their complexity and intra‑category diversity.
Low‑complexity categories are removed, and low‑diversity categories undergo reduced sampling.
To further enforce fine‑grained discrimination within categories, we retain about 120 high‑frequency and similar products within each sub‑category.
This procedure yields a high‑quality subset of approximately 20K products, which we refer to as \textbf{Catalog‑Fine}. To broaden the coverage of the catalog, we further sample an additional 1.3 M products from the remaining pool of eligible items, denoted as \textbf{Catalog‑Full}.

\textbf{\textit{Task Construction.}}
To instantiate tasks in the environment, we construct a set of shopping tasks in which each natural language purchase instruction pairs with a unique product from the entire catalog, ensuring unambiguous evaluation based on a single gold target.
To achieve this, we start with product metadata (title, category, store, options, attributes, and price) and ask annotators to write the instruction. To maintain naturalness and realism, they are required to avoid directly copying product titles or core keywords; instead, they employ synonym substitution, descriptive phrases, stylistic expressions, or everyday language to refer to product characteristics indirectly (e.g., a thoughtful gift, ``lightweight'' rather than a specific quantity). Finally, annotators verify that no other similar product fully satisfies the instruction. 
Based on this one-to-one mapping between instructions and real products, we create 24K tasks for Catalog-Fine. 
It should be noted that only in the single‑turn, non‑personalized setting, the user input is exactly the instruction, whereas in the multi‑turn setting, the goal is jointly expressed by the user and the agent through multiple interactions.

\textbf{\textit{Personalization.}}
To incorporate personalization, we augment each task with a structured user profile encoding both user information and long‑term preference cues (e.g., age, gender, spending tier, brand affinities).
Meanwhile, we remove these details from the instruction itself, retaining only product‑specific cues (e.g., color choice). 
This prevents the agent from taking a shortcut by inferring the complete requirement solely from the dialogue.
Due to the privacy constraints of real user profile, we start from existing product–instruction pairs and employ an LLM to generate an initial profile draft, embedding the latent long‑term needs implied in the instruction naturally into user attributes. Human annotators then review and revise these drafts to enrich the long-term preferences of user profiles so that they reflect real-world scenarios while avoiding overfitting to target product; overly specific features and requirements are kept in the instruction.
In total, we constructed 4,726 personalized user profiles with corresponding revised instructions.

\textbf{\textit{Multi-turn Interaction.}}
Compared to providing complete and explicit goals, user intentions in the real world are often initially incomplete, gradually clarified through ongoing interactions, and frequently expressed in an indirect manner.
To capture these characteristics, we extend ShopSimulator to support multi‑turn dialogues. 
We employ a LLM to simulate a shopper who begins with a deliberately vague target and reveals the necessary attributes only in response to the agent’s clarification requests. This setting requires the agent to engage in interactive disambiguation, actively identifying and filling in missing information before making a purchase, thereby aligning more closely with authentic user interaction.
The  role-playing prompt is shown in Figure \ref{fig:user_system_prompt}.

\subsection{Rewarding}
\label{subsec:reward}
A direct task reward design is to assign \(1\) if the agent fully satisfies all requirements and \(0\) otherwise. However, such a binary scheme makes it difficult to learn in stages, as partial fulfillment of the user’s request is not rewarded. 

\textbf{\textit{Loose (Additive) Reward.}}
First, following the reward design in WebShop (denoted as \(R_{\text{loose}}\)), at the end of an episode, if the agent purchases a product \(y\) with attribute set \(Y_{\text{att}}\), option set \(Y_{\text{opt}}\), and price \(Y_{\text{price}}\), the reward is computed as:  
\begin{equation}
\begin{aligned}
R_{\text{loose}}
= R_{\text{cat}}\cdot
\frac{
\begin{aligned}
&\left| U_{\text{att}}\!\cap\!Y_{\text{att}} \right|
+ \left| U_{\text{opt}}\!\cap\!Y_{\text{opt}} \right| \\
&\quad + \mathbf{1}\!\left[ Y_{\text{price}} \le U_{\text{price}} \right]
\end{aligned}
}{
\left| U_{\text{att}} \right|
+ \left| U_{\text{opt}} \right|
+ 1
}
\in[0,1]
\end{aligned}
\end{equation}
, where \(U_{\text{cat}},U_{\text{att}},U_{\text{opt}},U_{\text{price}}\) denote the target product’s category, attributes, option, and price constraint, respectively. 
\(R_{\text{cat}}\) is a soft match score that takes into account the similarity between \(Y_{\text{cat}}\) and \(U_{\text{cat}}\)\footnote{See Appendix \ref{subapp:reward_details} for detailed calculation introduction.}.
Under the condition of satisfying the category constraint, the reward is calculated as the average satisfaction of attributes, options and price.

\textbf{\textit{Strict (Multiplicative) Reward.}}
 In contrast, we consider a stricter multiplicative variant, denoted as \(R_{\text{strict}}\), which applies a bottleneck principle: the overall reward sharply decreases if any single constraint is not met. Formally,  
 \begin{equation}
\begin{aligned}
R_{\text{strict}}
&= r_{\text{cat}} \cdot
\frac{\left| U_{\text{att}} \cap Y_{\text{att}} \right|}{\left| U_{\text{att}} \right|}
\cdot
\frac{\left| U_{\text{opt}} \cap Y_{\text{opt}} \right|}{\left| U_{\text{opt}} \right|} \\
&\quad \cdot
\mathbf{1}\!\left[ Y_{\text{price}} \le U_{\text{price}} \right] \in [0,1]
\end{aligned}
\end{equation}

\subsection{Statistics}
\label{subsec:stastistic}

Figure~\ref{fig:env_info} shows the statistics of ShopSimulator.
In terms of tasks, based on the number of agent–user dialogue turns and whether personalization is involved, ShopSimulator covers four scenarios:
(1) \textbf{Single-Turn} (2) \textbf{Multi-Turn} (3) Single-Turn with personalization (\textbf{Single Turn \& Pers}) (4) Multi-Turn with personalization (\textbf{Multi-Turn \& Pers}). These 28K tasks are further split into evaluation and training sets.
Furthermore, Catalog-Fine contains 24K carefully curated products, fully covering the target items required in all tasks. Each product is annotated with attributes and several optional features (e.g., size, color) to support diverse dialogue and shopping scenarios.
Due to the resource consumption and latency introduced by the large product catalog, we use Catalog‑Fine for all subsequent evaluation and training experiments.

\begin{table*}[!t]
\small
\setlength{\tabcolsep}{1.0mm}{
    \begin{adjustbox}{width=\textwidth, keepaspectratio}
    \begin{tabular}{lccccccccccccccc}
        \toprule
        \multirow{2}{*}{\textbf{Model}} & \multicolumn{3}{c}{\textbf{Single Turn}} & \multicolumn{3}{c}{\textbf{Single Turn \& Pers.}}  & \multicolumn{3}{c}{\textbf{Multi-Turn}} & \multicolumn{3}{c}{\textbf{Multi-Turn \& Pers.}}  & \multicolumn{3}{c}{\textbf{Overall}} \\ 
        \cmidrule(lr){2-4}\cmidrule(lr){5-7}\cmidrule(lr){8-10}\cmidrule(lr){11-13}\cmidrule(lr){14-16}

          & $R_{\text{loose}}$ & $R_{\text{strict}}$ & $R_{\text{succ}}$ & $R_{\text{loose}}$ & $R_{\text{strict}}$ & $R_{\text{succ}}$ & $R_{\text{loose}}$ & $R_{\text{strict}}$ & $R_{\text{succ}}$ & $R_{\text{loose}}$ & $R_{\text{strict}}$ & $R_{\text{succ}}$ & $R_{\text{loose}}$ & $R_{\text{strict}}$ & $R_{\text{succ}}$ \\ 
        \midrule
        \multicolumn{16}{l}{\textit{\textbf{Closed-Source LLM}}} \\ 
        GPT-5~\citep{GPT-5} & \highgreen{63.88} & \highgreen{44.79} & \highgreen{40.78} & \highgreen{64.21} & \highgreen{41.19} & \highgreen{36.41} & 54.85 & 32.68 & 29.29 & 54.91 & 27.68 & 24.13 & 59.46 & \highgreen{36.58} & \highgreen{32.65} \\ 
        OpenAI-o3 \citep{OpenAI-o3} & 58.71 & 37.94 & 33.72 & 58.22 & 31.09 & 26.73 & 56.39 & 35.08 & 31.19 & 53.57 & 25.59 & 22.31 & 56.72 & 32.42 & 28.48 \\ 
        GPT-4.1~\citep{GPT-4.1} & 58.90 & 33.68 & 29.13 & 58.12 & 31.96 & 27.85 & 45.71 & 25.85 & 23.17 & 56.55 & 27.80 & 23.83 & 54.82 & 29.82 & 25.99 \\ 
        Claude-4-Sonnet~\citep{Claude-4} & 61.24 & 34.63 & 30.29 & 61.66 & 36.23 & 32.17 & \highgreen{64.96} & \highgreen{42.64} & \highgreen{38.25} & 57.97 & 29.08 & 25.39 & \highgreen{61.46} & 35.65 & 31.52 \\ 
        Gemini 2.5 Pro~\citep{comanici2025gemini} & 60.31 & 33.27 & 29.24 & 60.90 & 36.60 & 32.76 & 44.25 & 23.75 & 29.10 & \highgreen{58.40} & \highgreen{31.41} & \highgreen{27.13} & 55.97 & 31.26 & 29.55 \\ 
        \midrule
        \multicolumn{16}{l}{\textit{\textbf{Open-Source LLM}}} \\ 
        DeepSeek-R1 (671B-A37B) \citep{guo2025deepseek} & 50.95 & 29.78 & 25.27 & 50.73 & 32.64 & 29.19 & \highblue{58.39} & 32.99 & 29.27 & \highblue{63.09} & \highblue{34.05} & \highblue{30.3} & 55.79 & 32.37& 28.51 \\ 
        DeepSeek-V3.1 (671B-A37B)~\citep{DeepSeek-v3.1} & \highblue{63.76} & 37.42 & \highblue{31.86} & \highblue{65.86} & \highblue{40.31} & \highblue{35.69} & 54.78 & \highblue{35.54} & \highblue{31.62} & 59.13 & 32.39 & 28.07 & \highblue{60.89} & \highblue{36.42} & \highblue{31.81} \\ 
        Kimi-K2 (1T-A32B) \citep{team2025kimi} & 55.14 & 30.85 & 27.21 & 58.22 & 30.09 & 26.73 & 50.55 & 22.78 & 19.47 & 58.94 & 27.81 & 23.83 & 55.71 & 27.88 & 24.31 \\ 
        GLM-4.5 (335B-A32B) \citep{zeng2025glm} & 53.38 & 31.96 & 28.08 & 58.08 & 33.08 & 28.85 & 34.32 & 27.47 & 21.14 & 52.97 & 26.91 & 23.38 & 49.69 & 29.86 & 25.36 \\ 
        GPT-OSS-120B (-A5.1B) \citep{gpt-oss} & 61.82 & \highblue{39.13} & 21.42 & 44.67 & 20.80 & 17.48 & 43.87 & 20.33 & 16.78 & 47.71 & 15.96 & 12.29 & 49.52 & 24.06 & 16.99 \\ 
        Qwen3-235B-A22B \citep{yang2025qwen3} & 59.61 & 32.31 & 27.96 & 59.77 & 31.74 & 26.81 & 51.54 & 28.17 & 24.29 & 57.48 & 28.03 & 23.60 & 57.10 & 30.06 & 25.66 \\ 
        Qwen3-30B-A3B \citep{yang2025qwen3} & 56.69 & 25.74 & 20.88 & 61.48 & 28.22 & 22.84 & 48.07 & 12.70 & 10.04 & 53.43 & 21.05 & 17.18 & 54.92 & 21.93 & 17.73 \\ 
        Qwen3-32B \citep{yang2025qwen3} & 46.71 & 24.02 & 19.67 & 54.91 & 28.17 & 23.99 & 26.37 & 4.87 & 3.67 & 51.59 & 17.82 & 14.36 & 44.89 & 18.72 & 15.42 \\ 
        Qwen3-8B \citep{yang2025qwen3} & 38.90 & 16.89 & 14.13 & 44.91 & 20.79 & 17.24 & 34.98 & 8.12 & 6.48 & 46.04 & 17.06 & 13.63 & 41.21 & 15.72 & 12.87 \\
        \bottomrule
    \end{tabular}
    \end{adjustbox}
    }
\caption{The performance (\%) of models in four scenarios. ``Pers'' is an abbreviation for ``personalization''. 
The ``Overall'' score is the average of the scores from the four scenarios. The highest scores for open-source and closed-source LLMs is marked in \textcolor[RGB]{0,119,51}{green} and \textcolor{blue}{blue} respectively.}
    \label{tab:main_evaluation}
\end{table*}

\section{Evaluation}
\label{sec:eval}

\subsection{Evaluation Settings}
\label{subsec:eval_setting}

 In addition to the loose reward  (\(R_{\text{loose}}\)) and strict reward  (\(R_{\text{strict}}\)) in Section \ref{subsec:reward}, we also use the success rate (\(R_{\text{succ}}\)): the proportion of episodes in which the purchased product exactly matches the target item in all required dimensions (Only 0 and 1 for an episode).
 We set the agent’s maximum action step count to 30 for single-turn scenarios and 40 for multi-turn scenarios.  
For multi-turn scenarios, the shopper is played by Qwen3-225B-A22B. The results are shown in Table \ref{tab:main_evaluation}.


\subsection{Evaluation Results}
\label{subsec:eval_result}

\begin{figure}[t]
    \centering
\includegraphics[width=0.47\textwidth]{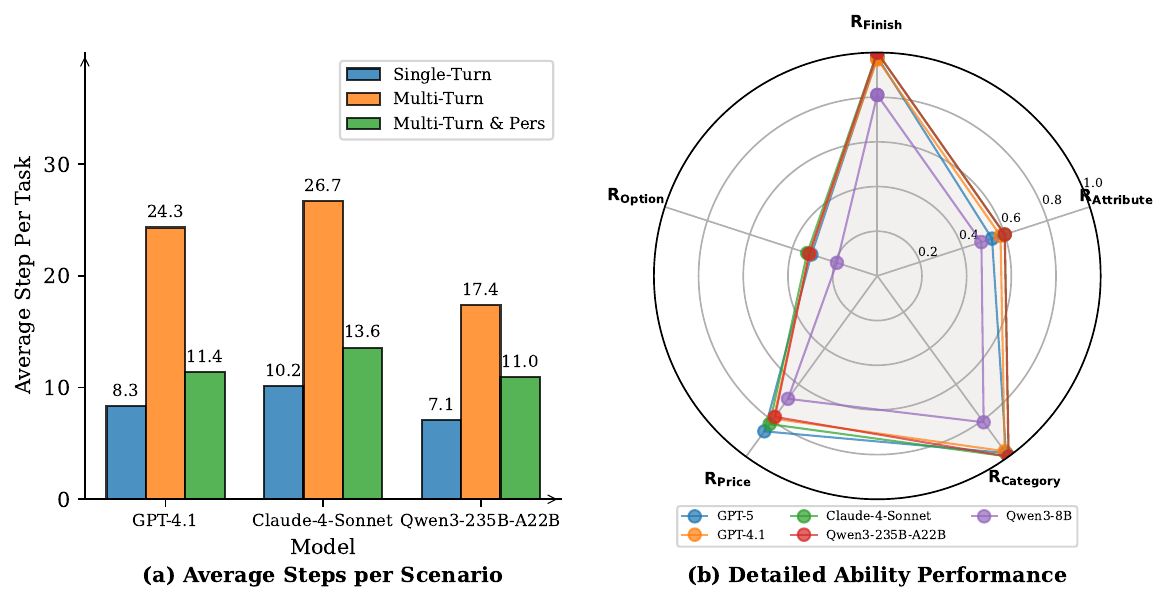} 
\caption{(a) shows agents’ average action steps per task in different scenarios. (b) shows detailed performance, where $R_{\text{finish}}$ indicates whether the agent finally recommends a product, regardless of accuracy.}
    \label{fig:eval_analysis}
\end{figure}

\textbf{\textit{Overall.}}
First, all task success rates are generally low, indicating that ShopSimulator presents a significant challenge to current LLM Agents. In terms of overall scores, even the best‑performing model, GPT‑5, achieves under 35\% on $R_{\text{succ}}$.
Second, LLMs consistently exhibit performance drops in multi‑turn and personalized scenarios, confirming the high difficulty posed by combining multi‑turn dialogue with personalization. 
Finally, the gap between loose, strict, and full‑success metrics is substantial. All models score much higher on $R_{\text{loose}}$ than on $R_{\text{strict}}$ and $R_{\text{succ}}$, suggesting that many recommendations match the main category or some attributes but fail to meet all fine‑grained requirements, as shown in Figure \ref{fig:eval_analysis}(b).

\textbf{\textit{Multi-Turn.}}
LLMs show a consistent score drop in multi-turn scenarios, indicating challenges in tracking in-context, clarifying, and understanding evolving intents over extended interactions. 
This effect is more pronounced in smaller models: for example, Qwen3‑8B’s strict score and success rate drop by 47\% and 50\% from single-turn to multi-turn, compared to Qwen3‑235B’s 13\% and 14\%. When comparing personalized single-turn with non-personalized multi-turn, most models (especially smaller ones) perform better in the former; for instance, Qwen3‑8B scores 12\% higher in strict and 11\% higher in success, suggesting that sustained clarification is more challenging for current LLMs than interpreting static preference constraints.

\textbf{\textit{Personalization.}}
The results indicate that the effect of introducing personalization is not monotonic across models. 
For example, under the Single‑Turn setting, Claude-4-Sonnet and Qwen3‑8B show gains, whereas GPT‑4.1 and Qwen3‑235B register slight drops. 
On the one hand, explicit user preferences help the agent narrow the search space early, reducing the need for repeated clarifications and unnecessary exploration. 
As shown in Figure \ref{fig:eval_analysis} (a), compared with Multi-Turn, providing LLMs with personalization information significantly reduces the number of actions required (including dialogue with the user).
On the other hand, redundant information in the user profile may be interpreted by the agent as additional hard constraints and increases contextual noise.

\begin{figure*}[t]
    \centering
    \resizebox{\textwidth}{!}{
    \includegraphics{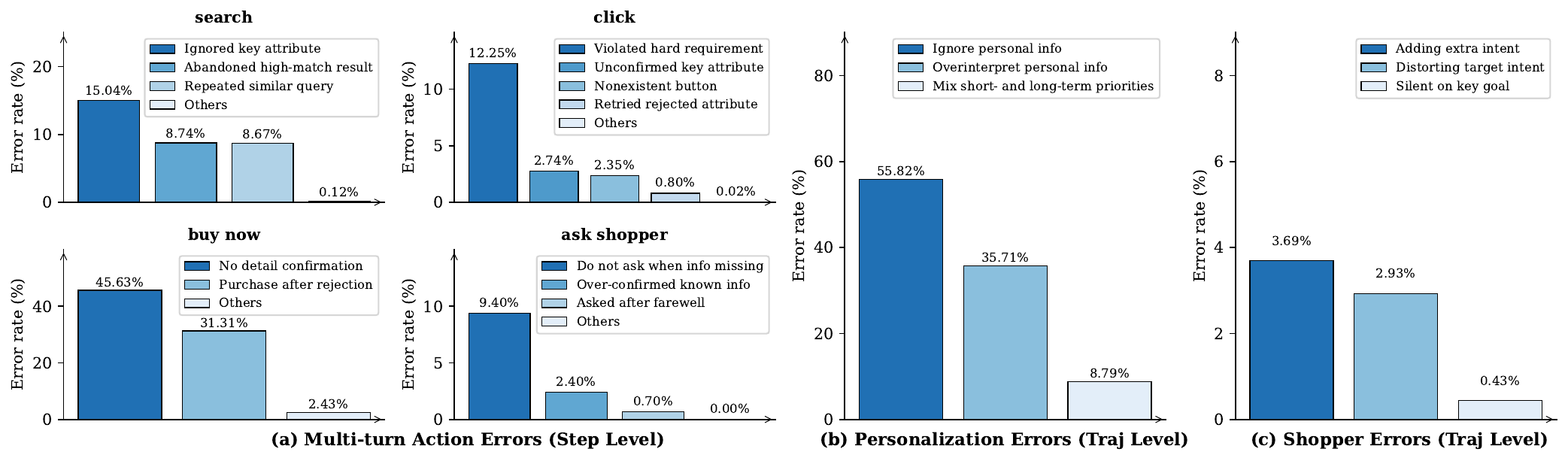}
    }
\caption{Error statistics of failed trajectories based on Claude‑4‑Sonnet: (a) Attribute errors for each action step of LLM under Multi‑Turn setting; (b) Categorize trajectory errors in terms of personalization under Single-Turn \& Pers setting; (c) Assess errors of the LLM-simulated Shopper. The introduction of errors is shown in Table \ref{tab:error_type_description}.}
    \label{fig:error_analysis}
\end{figure*}

\subsection{Error Analysis}
\label{subsec:error_analysis}

To investigate failure causes, we analyze error trajectories from Claude‑4‑Sonnet. We manually derive error types from a small set of tasks, then use GPT‑5 to classify errors for all trajectories.

\textbf{\textit{Action.}}
As shown in Figure \ref{fig:error_analysis}(a),
most errors in Search stem from insufficient use of available information, including ignoring key attributes, abandoning high‑match results, and issuing repeated queries, revealing weaknesses in state memory and retrieval consistency.
For Click, the dominant error is violating explicit user constraints, indicating inadequate enforcement and verification of conditions during execution.
For BuyNow, error rates are extremely high, with nearly 80\% of errors resulting from purchases without confirming details or despite explicit user rejection, indicating a lack of caution before finalizing decisions.
For AskShopper, the main issue is failing to inquire when information is missing (9.40\%), reflecting weak initiative in closing critical information gaps.
\textbf{Overall, the agent tends to ignore key attributes and underutilize existing results in retrieval, weakly enforce constraints in product selection, and make hasty purchase recommendations without sufficient user interaction, ultimately lowering task success rates and prolonging trajectories.}

\textbf{\textit{Personalization.}}
As shown in Figure \ref{fig:error_analysis}(b),
it reveals a strong concentration in two opposing issues: (1) partial omission of personalization information (55.82\%), where known user preference features are not fully utilized; and (2) over interpretation of personalization information (35.71\%), where limited preference data are extrapolated into overly specific or strict constraints, unreasonably narrowing the search space. In addition, confusion between immediate needs and long‑term preferences (8.79\%) leads to candidate rankings misaligned with the user’s actual intent.
\textbf{Overall, personalization errors mainly stem from the agent’s imbalance between underusing and overconstraining preference information.}

\textbf{\textit{Shopper.}}
We analyze failures caused by LLM‑simulated shoppers. As shown in Figure \ref{fig:error_analysis}(c), shopper‑side errors account for a relatively small proportion (less than 8\%). This suggests that agent‑side failures still remain the primary reason for the overall low task success rate.

\section{RL Exploration}
\label{sec:train}

\begin{table*}[!t]
    \centering
    \small
\setlength{\tabcolsep}{2.7mm}{
    \begin{adjustbox}{width=\textwidth, keepaspectratio}
    \begin{tabular}{lccccccccc}
    \toprule
        \multirow{2}{*}{\textbf{Scenario}} & \multirow{2}{*}{\textbf{Training}} & \multicolumn{3}{c}{\textbf{Metrics}} & \multicolumn{5}{c}{\textbf{Details}} \\ 
        \cmidrule(lr){3-5}\cmidrule(lr){6-10}
        &&  $R_{\text{loose}}$ & $R_{\text{strict}}$ & $R_{\text{succ}}$ & $R_{\text{finish}}$ & $R_{\text{category}}$ & $R_{\text{attribute}}$ & $R_{\text{option}}$ & $R_{\text{price}}$ \\ 
        \midrule
        \multirow{5}{*}{Single-Turn} & Baseline & 38.90 & 16.89 & 14.13 & 60.65 & 60.65 & 40.53 & 18.72 & 56.10 \\ 
        ~ & SFT & $62.72_{\textcolor{deepgreen}{(+23.82)}}$ & $37.22_{\textcolor{deepgreen}{(+20.33)}}$ & $32.47_{\textcolor{deepgreen}{(+18.34)}}$ & 94.88 & 94.88 & 63.41 & 41.92 & 84.29 \\ 
        ~ & RL w. $R_{\text{loose}}$ & $59.34_{\textcolor{deepgreen}{(+20.44)}}$ & $32.81_{\textcolor{deepgreen}{(+15.92)}}$  & $28.07_{\textcolor{deepgreen}{(+13.94)}}$ & 91.59 & 91.59 & 60.50 & 36.35 & 82.80  \\ 
        ~ & RL w. $R_{\text{strict}}$ & $62.52_{\textcolor{deepgreen}{(+23.62)}}$ & $34.85_{\textcolor{deepgreen}{(+17.96)}}$ & $30.19_{\textcolor{deepgreen}{(+16.06)}}$ & 94.10 &  94.10 & 63.96  & 38.39 & 85.50  \\ 
        ~ & SFT + RL w. $R_{\text{strict}}$ & $67.01_{\textcolor{deepgreen}{(+28.11)}}$ & $43.21_{\textcolor{deepgreen}{(+26.32)}}$ & $38.89_{\textcolor{deepgreen}{(+24.76)}}$ & 96.79 & 96.79 & 68.27 & 47.26 & 85.99  \\ 
        \midrule
        \multirow{5}{*}{Multi-Turn} & Baseline & 34.98 & 8.12 & 6.48 & 68.73 & 68.73 & 37.08 & 9.43 & 59.49 \\ 
        ~ & SFT & $45.60_{\textcolor{deepgreen}{(+10.62)}}$ & $32.54_{\textcolor{deepgreen}{(+24.42)}}$ & $29.27_{\textcolor{deepgreen}{(+22.79)}}$ & 62.17  & 62.17 & 45.88 & 35.46 & 56.68 \\ 
        ~ & RL w. $R_{\text{loose}}$ & $49.55_{\textcolor{deepgreen}{(+14.57)}}$ & $21.15_{\textcolor{deepgreen}{(+13.03)}}$ & $18.64_{\textcolor{deepgreen}{(+12.16)}}$ & 92.60 & 92.60 & 50.25 & 22.16 & 80.60 \\ 
        ~ & RL w. $R_{\text{strict}}$ & $53.89_{\textcolor{deepgreen}{(+18.98)}}$ & $24.32_{\textcolor{deepgreen}{(+16.20)}}$ & $22.32_{\textcolor{deepgreen}{(+15.84)}}$ & 97.46 & 97.46 & 54.61 & 25.67 & 86.29 \\ 
        ~ & SFT + RL w. $R_{\text{strict}}$ & $64.51_{\textcolor{deepgreen}{(+29.53)}}$ & $39.34_{\textcolor{deepgreen}{(+31.22)}}$ & $35.50_{\textcolor{deepgreen}{(29.02)}}$ & 98.11 & 98.11 & 66.02 & 43.05 & 84.63 \\ 
        \midrule
        \multirow{5}{*}{\makecell[c]{Single-Turn\\ \& Pers}} & Baseline & 44.91 & 20.79 & 17.24 & 70.67 & 70.67 & 46.56 & 23.16 & 65.22 \\ 
        ~ & SFT & $61.22_{\textcolor{deepgreen}{(+16.31)}}$ & $35.46_{\textcolor{deepgreen}{(+14.67)}}$ & $30.98_{\textcolor{deepgreen}{(+13.74)}}$ & 97.74 & 97.74 & 60.93 & 40.55 & 87.22  \\ 
        ~ & RL w. $R_{\text{loose}}$ & $66.91_{\textcolor{deepgreen}{(+22.00)}}$ & $38.66_{\textcolor{deepgreen}{(+17.87)}}$ & $32.98_{\textcolor{deepgreen}{(+15.74)}}$ & 99.77 & 99.77 & 69.48 & 42.22 & 88.32  \\ 
        ~ & RL w. $R_{\text{strict}}$ & $69.07_{\textcolor{deepgreen}{(+24.16)}}$ & $43.71_{\textcolor{deepgreen}{(+22.92)}}$ & $38.37_{\textcolor{deepgreen}{(+21.13)}}$ & 97.29 & 97.29 & 71.87 & 47.07 & 87.06 \\ 
        ~ & SFT + RL w. $R_{\text{strict}}$ & $71.48_{\textcolor{deepgreen}{(+26.57)}}$ & $60.18_{\textcolor{deepgreen}{(+39.39)}}$ & $57.33_{\textcolor{deepgreen}{(+40.09)}}$ & 83.08 & 83.08 & 72.23 & 63.12 & 78.22 \\ 
        \midrule
        \multirow{5}{*}{\makecell[c]{Multi-Turn\\ \& Pers}} & Baseline & 46.04 & 17.06 & 13.63 & 81.09 & 81.09 & 48.61 & 19.37 & 68.58  \\ 
        ~ & SFT & $63.62_{\textcolor{deepgreen}{(+17.58)}}$ & $35.97_{\textcolor{deepgreen}{(+18.91)}}$ & $30.49_{\textcolor{deepgreen}{(+16.86)}}$ & 99.63 & 99.63 & 66.49 & 40.17 & 81.95 \\ 
        ~ & RL w. $R_{\text{loose}}$ & $63.20_{\textcolor{deepgreen}{(+17.16)}}$ & $33.01_{\textcolor{deepgreen}{(+15.95)}}$  & $28.44_{\textcolor{deepgreen}{(+14.81)}}$ & 99.85 & 99.85 & 67.00 & 36.11 & 82.73 \\ 
        ~ & RL w. $R_{\text{strict}}$ & $64.01_{\textcolor{deepgreen}{(+17.97)}}$ & $33.18_{\textcolor{deepgreen}{(+16.12)}}$ & $28.29_{\textcolor{deepgreen}{(+14.66)}}$ & 99.78 & 99.78 & 67.62 & 36.30 & 83.99 \\ 
        ~ & SFT + RL w. $R_{\text{strict}}$ & $65.06_{\textcolor{deepgreen}{(+19.02)}}$ & $38.50_{\textcolor{deepgreen}{(+21.44)}}$ & $34.35_{\textcolor{deepgreen}{(+20.72)}}$ & 99.01 & 99.01 & 68.08 & 42.62 & 80.49  \\ 
        \bottomrule
    \end{tabular}
    \end{adjustbox}
    }
    \caption{The performance of Qwen3-8B (baseline), SFT, RL, and combined SFT+RL training across scenarios.}
\label{tab:train_result_all}
\end{table*}

\subsection{Training Settings}
\label{subsec:train_setting}
We perform RL training separately for each scenario and consider two configurations: (1) \textbf{directly performs RL} on Qwen3‑8B; (2) \textbf{SFT on Qwen3‑8B for cold‑start, followed by RL.}
For SFT, we collect 6K successful trajectories from GPT‑4.1 on the training set.
For RL, we use the GRPO algorithm~\citep{shao2024deepseekmath} and experiment with both reward: \(R_{\text{loose}}\) and \(R_{\text{strict}}\).
The detailed implementation is shown in Appendix~\ref{subapp:rl_implement}.
\subsection{Training Results}
\label{subsec:train_result}

\begin{figure*}[t]
    \centering
    \resizebox{\textwidth}{!}{
    \includegraphics{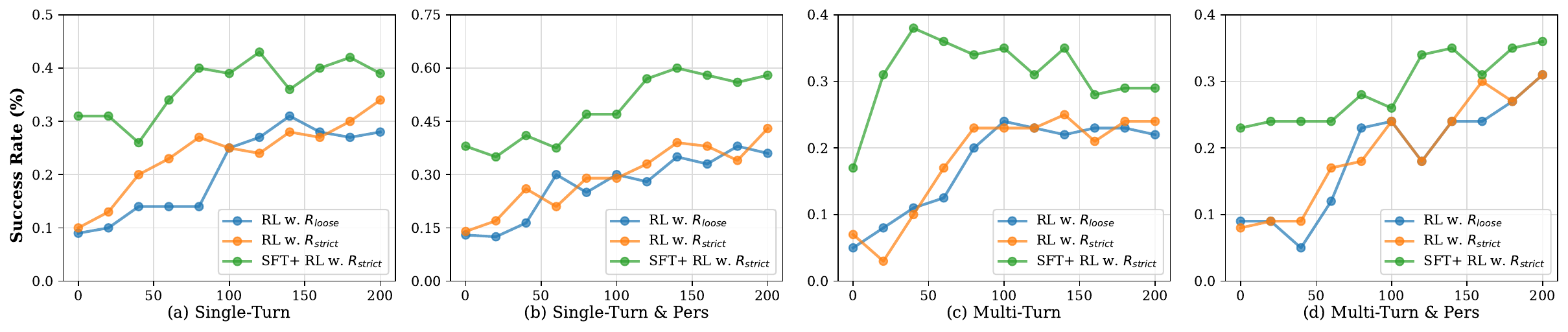}
    }
\caption{The performance of Qwen3-8B on the test set (random 128 samples) varies with RL training steps.}
    \label{fig:rl_curve}
\end{figure*}

\textbf{\textit{Overall.}}
As shown in Table \ref{tab:train_result_all} and Figure \ref{fig:rl_curve}, all training methods significantly outperform the baseline, with SFT + RL w. $R_{\text{strict}}$ achieving optimal results in all scenarios.
Metric breakdowns in Table~\ref{tab:train_result_all} show that these gains are mainly driven by improvements in attribute and option matching.

\begin{figure}[t]
    \centering
\includegraphics[width=0.47\textwidth]{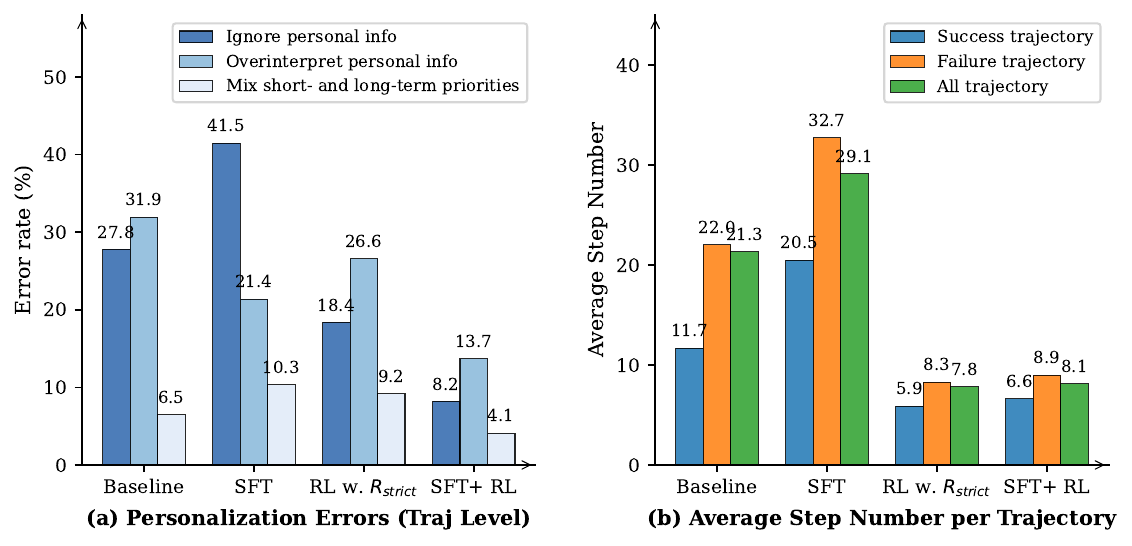} 
\caption{Fig (a) shows the error rate across all trajectories in  Single-Turn \& Pers, with successful trajectories assumed to have no errors. Fig (b) shows the average steps per trajectory in Multi-Turn.}
    \label{fig:train_analysis}
\end{figure}

\textbf{\textit{Learning Mode Comparison.}}
Comparison between SFT and standalone RL w. \(R_{strict}\):  
(1) \textbf{Without priors, RL has difficulty in autonomous exploration, especially in multi-turn tasks}; in contrast, SFT directly imitates successful trajectories, making learning easier. Consequently, RL lags behind SFT in both Single-Turn and Multi-Turn settings, with a gap of up to 6.95\% in the Multi-Turn scenario. 
Besides, as shown in Figure \ref{fig:train_analysis}(b), SFT learns the teacher model’s long successful trajectories, which leads to its own trajectories being similarly lengthy, whereas RL, through self-exploration, produces short and concise trajectories. 
(2) \textbf{SFT primarily learns the task workflow and patterns but struggles to capture preferences, whereas RL, through multiple rollouts, more easily identifies and optimizes personalized preferences}. Evidence comes from the fact that, after introducing personalization, the gap between RL and SFT narrows from 6.95\% to 2.79\%, and RL even surpasses SFT by +8.25\% in the single-turn personalized setting.
Furthermore, as shown in Figure \ref{fig:train_analysis} (a), SFT does not lead to a significant reduction in personalization errors like RL.
This also explains why using SFT for cold start followed by RL training yields significant improvements: On the one hand, SFT provides high-quality priors, reducing subsequent RL exploration difficulty. On the other hand, SFT first learns process patterns, while RL then learns preferences and improve weaknesses such as attribute and option matching.

\textbf{\textit{Reward Comparison.}}
Using \(R_{strict}\) as the RL training objective consistently outperforms using \(R_{loose}\) across all four scenarios. 
This can be attributed to \(R_{strict}\) being a typical bottleneck-style rewarding, which focuses optimization on the weakest dimensions (e.g., attributes, options), thereby achieving higher overall matching and success rates in most scenarios (see Figure \ref{tab:train_result_all}). 
In contrast, \(R_{loose}\)-trained models tend to optimize for coverage or task completion, with less emphasis on accuracy.  

\textbf{\textit{Scenario Comparison.}}
Single-Turn and Single-Turn \& Pers scenarios yield larger improvements through training, mainly due to the model learning a complete task execution flow and effectively leveraging user profiles.
In contrast, Multi-Turn and Multi-Turn \& Pers scenarios require the agent to maintain long-term consistency and coordination across personalization understanding, clarification dialogue, and environment actions, markedly increasing complexity; even after training, success rates remain around 35\%.

\section{Related Work}
\label{sec:related_work}

\subsection{Evaluating LLM Agent for E-commerce}
\label{subsec:evaluate_related_work}

Recently, LLMs have attracted increasing attention in e‑commerce~\citep{xu2024survey}. 
Several works focus on evaluating LLMs' fundamental e‑commerce concepts and knowledge reasoning~\citep{jin2024shopping,chen2025chineseecomqa,liu2025eckgbench}. 
Other studies emphasize assessing LLMs' ability to simulate user actions in e‑commerce ~\citep{wang2025opera,sun2025llm}.
In this paper, we focus on evaluating LLMs' ability to integratively complete shopping tasks within an environment. WebShop~\citep{yao2022webshop} pioneered the construction of shopping environments, while ChatShop, DeepShop, WebMall, and ShoppingBench~\citep{chen2024chatshopinteractiveinformationseeking,lyu2025deepshopbenchmarkdeepresearch,peeters2025webmallmultishopbenchmark,wang2025shoppingbenchrealworldintentgroundedshopping} extended it to multi‑turn dialogue, deep product search, cross‑store browsing, and voucher usage, respectively.
However, they primarily rely on English product corpora and lack personalization and training support. In contrast, ShopSimulator provides a large‑scale Chinese e‑commerce sandbox to evaluate whether LLMs can serve as reliable shopping assistants in settings with personalized user preferences and multi‑turn interactions.

\subsection{Developing LLM Agents for E-commerce}
\label{subsec:training_related_work}
Some studies aim to enhance LLMs’ fundamental e-commerce capabilities~\citep{zhang2024llasa,herold2025domain,li2024ecomgpt}.
We mainly discuss the development of LLM agent for shopping assistants.
WebShop~\citep{yao2022webshop} first train a shopping agent using a BERT backbone; 
Hybrid‑MACRS~\citep{nie2024hybrid} builds a conversational recommendation system via multi‑agent collaboration;
MindFlow~\citep{gong2025mindflow} targets multimodality; 
RecGPT~\citep{yi2025recgpt} integrates LLMs into the recommendation pipeline; 
Shop‑R1~\citep{zhang2025shop} leverages RL to better simulate real human shopping behavior; 
and ShoppingBench~\citep{lyu2025deepshopbenchmarkdeepresearch} combines SFT for long-horizon interaction learning with RL to improve tool-use.
In this paper, we first build ShopSimulator to provide abundant training tasks and multi‑dimensional reward signals; and through systematic experiments, we compare RL with SFT, and examine the effect of SFT cold‑start on RL.

\section{Conclusion}
\label{sec:conclusion}
In this paper, we introduce ShopSimulator, a unified Chinese shopping sandbox environment that couples large‑scale real product catalogs with multi‑turn, personalized user modeling and provides both evaluation and training support.
Our evaluation reveals that even strong LLMs are far from reliable shopping assistant agents.
Most errors stem from failures in fine‑grained attribute/option grounding and suboptimal interaction behavior, and combining SFT with RL training can partially address these shortcomings.
Looking ahead, we expect ShopSimulator to serve as a reliable platform to advance LLM agents for shopping assistants.


\section*{Limitations}
In this paper, we propose a ShopSimulator to support the evaluation and training of LLM agents in e-commerce shopping scenarios.
However, there are still some limitations as follows:
(1) Due to privacy, the personalized user profiles in ShopSimulator are synthetically generated by LLMs and subsequently reviewed and refined by human annotators. As a result, their distribution may deviate from that of real-world user data.
(2) In our RL experiments, we explored diverse scenarios and settings, but the policy optimization was primarily conducted using GRPO; recent advances in RL algorithms were not investigated.
(3) This work focuses on text-based interaction, without incorporating the multimodal data, such as product images and videos in realistic e-commerce recommendation systems.

\bibliography{custom}

\begin{thebibliography}{33}
\providecommand{\natexlab}[1]{#1}

\bibitem[{Chen et~al.(2025)Chen, Lv, Hu, Li, Yuan, He, Zhang, Liu, Liu, Su
  et~al.}]{chen2025chineseecomqa}
Haibin Chen, Kangtao Lv, Chengwei Hu, Yanshi Li, Yujin Yuan, Yancheng He,
  Xingyao Zhang, Langming Liu, Shilei Liu, Wenbo Su, and 1 others. 2025.
\newblock Chineseecomqa: A scalable e-commerce concept evaluation benchmark for
  large language models.
\newblock In \emph{Proceedings of the 31st ACM SIGKDD Conference on Knowledge
  Discovery and Data Mining V. 2}, pages 5311--5321.

\bibitem[{Chen et~al.(2024)Chen, Wiseman, and
  Dhingra}]{chen2024chatshopinteractiveinformationseeking}
Sanxing Chen, Sam Wiseman, and Bhuwan Dhingra. 2024.
\newblock \href {https://arxiv.org/abs/2404.09911} {Chatshop: Interactive
  information seeking with language agents}.
\newblock \emph{Preprint}, arXiv:2404.09911.

\bibitem[{Claude(2025)}]{Claude-4}
Claude. 2025.
\newblock \href {https://www.anthropic.com/news/claude-4} {Introducing claude 4
  | anthropic}.

\bibitem[{Comanici et~al.(2025)Comanici, Bieber, Schaekermann, Pasupat,
  Sachdeva, Dhillon, Blistein, Ram, Zhang, Rosen et~al.}]{comanici2025gemini}
Gheorghe Comanici, Eric Bieber, Mike Schaekermann, Ice Pasupat, Noveen
  Sachdeva, Inderjit Dhillon, Marcel Blistein, Ori Ram, Dan Zhang, Evan Rosen,
  and 1 others. 2025.
\newblock Gemini 2.5: Pushing the frontier with advanced reasoning,
  multimodality, long context, and next generation agentic capabilities.
\newblock \emph{arXiv preprint arXiv:2507.06261}.

\bibitem[{DeepSeek(2025)}]{DeepSeek-v3.1}
DeepSeek. 2025.
\newblock \href {https://api-docs.deepseek.com/news/news250821} {Deepseek-v3.1
  release | deepseek api docs}.

\bibitem[{Gong et~al.(2025)Gong, Huang, Yang, Peng, Wang, Liu, and
  Jiang}]{gong2025mindflow}
Ming Gong, Xucheng Huang, Chenghan Yang, Xianhan Peng, Haoxin Wang, Yang Liu,
  and Ling Jiang. 2025.
\newblock Mindflow: Revolutionizing e-commerce customer support with multimodal
  llm agents.
\newblock \emph{arXiv preprint arXiv:2507.05330}.

\bibitem[{Guo et~al.(2025)Guo, Yang, Zhang, Song, Zhang, Xu, Zhu, Ma, Wang, Bi
  et~al.}]{guo2025deepseek}
Daya Guo, Dejian Yang, Haowei Zhang, Junxiao Song, Ruoyu Zhang, Runxin Xu,
  Qihao Zhu, Shirong Ma, Peiyi Wang, Xiao Bi, and 1 others. 2025.
\newblock Deepseek-r1: Incentivizing reasoning capability in llms via
  reinforcement learning.
\newblock \emph{arXiv preprint arXiv:2501.12948}.

\bibitem[{Herold et~al.(2025)Herold, Kozielski, Bazazo, Petrushkov, Cieplicka,
  Basaj, Versley, Hashemi, and Khadivi}]{herold2025domain}
Christian Herold, Michael Kozielski, Tala Bazazo, Pavel Petrushkov, Patrycja
  Cieplicka, Dominika Basaj, Yannick Versley, Seyyed~Hadi Hashemi, and Shahram
  Khadivi. 2025.
\newblock Domain adaptation of foundation llms for e-commerce.
\newblock \emph{arXiv preprint arXiv:2501.09706}.

\bibitem[{Jin et~al.(2024)Jin, Li, Zhang, Cao, Gao, Jayarao, Li, Liu, Sarkhel,
  Tang et~al.}]{jin2024shopping}
Yilun Jin, Zheng Li, Chenwei Zhang, Tianyu Cao, Yifan Gao, Pratik Jayarao, Mao
  Li, Xin Liu, Ritesh Sarkhel, Xianfeng Tang, and 1 others. 2024.
\newblock Shopping mmlu: A massive multi-task online shopping benchmark for
  large language models.
\newblock \emph{Advances in Neural Information Processing Systems},
  37:18062--18089.

\bibitem[{Li et~al.(2024)Li, Ma, Wang, Huang, Jiang, Zheng, Xie, Huang, and
  Jiang}]{li2024ecomgpt}
Yangning Li, Shirong Ma, Xiaobin Wang, Shen Huang, Chengyue Jiang, Hai-Tao
  Zheng, Pengjun Xie, Fei Huang, and Yong Jiang. 2024.
\newblock Ecomgpt: Instruction-tuning large language models with chain-of-task
  tasks for e-commerce.
\newblock In \emph{Proceedings of the AAAI Conference on Artificial
  Intelligence}, volume~38, pages 18582--18590.

\bibitem[{Liu et~al.(2025)Liu, Chen, Wang, Yuan, Liu, Su, Zhao, and
  Zheng}]{liu2025eckgbench}
Langming Liu, Haibin Chen, Yuhao Wang, Yujin Yuan, Shilei Liu, Wenbo Su,
  Xiangyu Zhao, and Bo~Zheng. 2025.
\newblock Eckgbench: Benchmarking large language models in e-commerce
  leveraging knowledge graph.
\newblock \emph{arXiv preprint arXiv:2503.15990}.

\bibitem[{Lyu et~al.(2025)Lyu, Zhang, Yan, de~Rijke, Ren, and
  Chen}]{lyu2025deepshopbenchmarkdeepresearch}
Yougang Lyu, Xiaoyu Zhang, Lingyong Yan, Maarten de~Rijke, Zhaochun Ren, and
  Xiuying Chen. 2025.
\newblock \href {https://arxiv.org/abs/2506.02839} {Deepshop: A benchmark for
  deep research shopping agents}.
\newblock \emph{Preprint}, arXiv:2506.02839.

\bibitem[{Nie et~al.(2024)Nie, Zhi, Yan, Du, Zhang, Chen, Zhou, Chen, Li, Cheng
  et~al.}]{nie2024hybrid}
Guangtao Nie, Rong Zhi, Xiaofan Yan, Yufan Du, Xiangyang Zhang, Jianwei Chen,
  Mi~Zhou, Hongshen Chen, Tianhao Li, Ziguang Cheng, and 1 others. 2024.
\newblock A hybrid multi-agent conversational recommender system with llm and
  search engine in e-commerce.
\newblock In \emph{Proceedings of the 18th ACM Conference on Recommender
  Systems}, pages 745--747.

\bibitem[{OpenAI(2025{\natexlab{a}})}]{GPT-5}
OpenAI. 2025{\natexlab{a}}.
\newblock \href {https://openai.com/gpt-5/} {Gpt-5 is here | openai}.

\bibitem[{OpenAI(2025{\natexlab{b}})}]{GPT-4.1}
OpenAI. 2025{\natexlab{b}}.
\newblock \href {https://openai.com/index/gpt-4-1/} {Introducing gpt-4.1 in the
  api | openai}.

\bibitem[{OpenAI(2025{\natexlab{c}})}]{gpt-oss}
OpenAI. 2025{\natexlab{c}}.
\newblock \href {https://openai.com/index/introducing-gpt-oss/} {Introducing
  gpt-oss | openai}.

\bibitem[{OpenAI(2025{\natexlab{d}})}]{OpenAI-o3}
OpenAI. 2025{\natexlab{d}}.
\newblock \href {https://openai.com/index/introducing-o3-and-o4-mini/}
  {Introducing openai o3 and o4-mini | openai}.

\bibitem[{Peeters et~al.(2025)Peeters, Steiner, Schwarz, Caspary, and
  Bizer}]{peeters2025webmallmultishopbenchmark}
Ralph Peeters, Aaron Steiner, Luca Schwarz, Julian~Yuya Caspary, and Christian
  Bizer. 2025.
\newblock \href {https://arxiv.org/abs/2508.13024} {Webmall -- a multi-shop
  benchmark for evaluating web agents}.
\newblock \emph{Preprint}, arXiv:2508.13024.

\bibitem[{Rafailov et~al.(2023)Rafailov, Sharma, Mitchell, Manning, Ermon, and
  Finn}]{rafailov2023direct}
Rafael Rafailov, Archit Sharma, Eric Mitchell, Christopher~D Manning, Stefano
  Ermon, and Chelsea Finn. 2023.
\newblock Direct preference optimization: Your language model is secretly a
  reward model.
\newblock \emph{Advances in neural information processing systems},
  36:53728--53741.

\bibitem[{Shao et~al.(2024)Shao, Wang, Zhu, Xu, Song, Bi, Zhang, Zhang, Li
  et~al.}]{shao2024deepseekmath}
Zhihong Shao, Peiyi Wang, Qihao Zhu, Runxin Xu, Junxiao Song, Xiao Bi, Haowei
  Zhang, Mingchuan Zhang, YK~Li, and 1 others. 2024.
\newblock Deepseekmath: Pushing the limits of mathematical reasoning in open
  language models.
\newblock \emph{arXiv preprint arXiv:2402.03300}.

\bibitem[{Sun et~al.(2025)Sun, Fu, Yao, Lu, Li, Gu, Gesi, Huang, Luo, and
  Wang}]{sun2025llm}
Lu~Sun, Shihan Fu, Bingsheng Yao, Yuxuan Lu, Wenbo Li, Hansu Gu, Jiri Gesi,
  Jing Huang, Chen Luo, and Dakuo Wang. 2025.
\newblock Llm agent meets agentic ai: Can llm agents simulate customers to
  evaluate agentic-ai-based shopping assistants?
\newblock \emph{arXiv preprint arXiv:2509.21501}.

\bibitem[{Team et~al.(2025)Team, Bai, Bao, Chen, Chen, Chen, Chen, Chen, Chen,
  Chen et~al.}]{team2025kimi}
Kimi Team, Yifan Bai, Yiping Bao, Guanduo Chen, Jiahao Chen, Ningxin Chen,
  Ruijue Chen, Yanru Chen, Yuankun Chen, Yutian Chen, and 1 others. 2025.
\newblock Kimi k2: Open agentic intelligence.
\newblock \emph{arXiv preprint arXiv:2507.20534}.

\bibitem[{Wang et~al.(2025{\natexlab{a}})Wang, Xiao, Sun, Zhao, Luo, Zhang, and
  Zeng}]{wang2025shoppingbenchrealworldintentgroundedshopping}
Jiangyuan Wang, Kejun Xiao, Qi~Sun, Huaipeng Zhao, Tao Luo, Jiandong Zhang, and
  Xiaoyi Zeng. 2025{\natexlab{a}}.
\newblock \href {https://arxiv.org/abs/2508.04266} {Shoppingbench: A real-world
  intent-grounded shopping benchmark for llm-based agents}.
\newblock \emph{Preprint}, arXiv:2508.04266.

\bibitem[{Wang et~al.(2025{\natexlab{b}})Wang, Xiong, Chen, Gao, Guo, He,
  Huang, Liu, Li, Li et~al.}]{wang2025reinforcement}
Weixun Wang, Shaopan Xiong, Gengru Chen, Wei Gao, Sheng Guo, Yancheng He,
  Ju~Huang, Jiaheng Liu, Zhendong Li, Xiaoyang Li, and 1 others.
  2025{\natexlab{b}}.
\newblock Reinforcement learning optimization for large-scale learning: An
  efficient and user-friendly scaling library.
\newblock \emph{arXiv preprint arXiv:2506.06122}.

\bibitem[{Wang et~al.(2025{\natexlab{c}})Wang, Lu, Li, Amini, Sun, Bart, Lyu,
  Gesi, Wang, Huang et~al.}]{wang2025opera}
Ziyi Wang, Yuxuan Lu, Wenbo Li, Amirali Amini, Bo~Sun, Yakov Bart, Weimin Lyu,
  Jiri Gesi, Tian Wang, Jing Huang, and 1 others. 2025{\natexlab{c}}.
\newblock Opera: A dataset of observation, persona, rationale, and action for
  evaluating llms on human online shopping behavior simulation.
\newblock \emph{arXiv preprint arXiv:2506.05606}.

\bibitem[{Wu et~al.(2024)Wu, Zheng, Qiu, Wang, Gu, Shen, Qin, Zhu, Zhu, Liu
  et~al.}]{wu2024survey}
Likang Wu, Zhi Zheng, Zhaopeng Qiu, Hao Wang, Hongchao Gu, Tingjia Shen, Chuan
  Qin, Chen Zhu, Hengshu Zhu, Qi~Liu, and 1 others. 2024.
\newblock A survey on large language models for recommendation.
\newblock \emph{World Wide Web}, 27(5):60.

\bibitem[{Xu et~al.(2024)Xu, Zhang, Yang, Yang, Xu, Zheng, and
  Liang}]{xu2024survey}
Da~Xu, Danqing Zhang, Guangyu Yang, Bo~Yang, Shuyuan Xu, Lingling Zheng, and
  Cindy Liang. 2024.
\newblock Survey for landing generative ai in social and e-commerce recsys--the
  industry perspectives.
\newblock \emph{arXiv preprint arXiv:2406.06475}.

\bibitem[{Yang et~al.(2025)Yang, Li, Yang, Zhang, Hui, Zheng, Yu, Gao, Huang,
  Lv et~al.}]{yang2025qwen3}
An~Yang, Anfeng Li, Baosong Yang, Beichen Zhang, Binyuan Hui, Bo~Zheng, Bowen
  Yu, Chang Gao, Chengen Huang, Chenxu Lv, and 1 others. 2025.
\newblock Qwen3 technical report.
\newblock \emph{arXiv preprint arXiv:2505.09388}.

\bibitem[{Yao et~al.(2022)Yao, Chen, Yang, and Narasimhan}]{yao2022webshop}
Shunyu Yao, Howard Chen, John Yang, and Karthik Narasimhan. 2022.
\newblock Webshop: Towards scalable real-world web interaction with grounded
  language agents.
\newblock \emph{Advances in Neural Information Processing Systems},
  35:20744--20757.

\bibitem[{Yi et~al.(2025)Yi, Chen, Guo, Tang, Wu, Yu, Zhang, Dai, Chen, Yang
  et~al.}]{yi2025recgpt}
Chao Yi, Dian Chen, Gaoyang Guo, Jiakai Tang, Jian Wu, Jing Yu, Mao Zhang,
  Sunhao Dai, Wen Chen, Wenjun Yang, and 1 others. 2025.
\newblock Recgpt technical report.
\newblock \emph{arXiv preprint arXiv:2507.22879}.

\bibitem[{Zeng et~al.(2025)Zeng, Lv, Zheng, Hou, Chen, Xie, Wang, Yin, Zeng,
  Zhang et~al.}]{zeng2025glm}
Aohan Zeng, Xin Lv, Qinkai Zheng, Zhenyu Hou, Bin Chen, Chengxing Xie, Cunxiang
  Wang, Da~Yin, Hao Zeng, Jiajie Zhang, and 1 others. 2025.
\newblock Glm-4.5: Agentic, reasoning, and coding (arc) foundation models.
\newblock \emph{arXiv preprint arXiv:2508.06471}.

\bibitem[{Zhang et~al.(2024)Zhang, Peng, Zhao, Hu, Zhu, Zeng, and
  Hu}]{zhang2024llasa}
Shuo Zhang, Boci Peng, Xinping Zhao, Boren Hu, Yun Zhu, Yanjia Zeng, and Xuming
  Hu. 2024.
\newblock Llasa: Large language and e-commerce shopping assistant.
\newblock In \emph{Amazon KDD Cup 2024 Workshop}.

\bibitem[{Zhang et~al.(2025)Zhang, Wang, Gesi, Wang, Lu, Lin, Zhan, Gao, Jiao,
  Liu et~al.}]{zhang2025shop}
Yimeng Zhang, Tian Wang, Jiri Gesi, Ziyi Wang, Yuxuan Lu, Jiacheng Lin, Sinong
  Zhan, Vianne Gao, Ruochen Jiao, Junze Liu, and 1 others. 2025.
\newblock Shop-r1: Rewarding llms to simulate human behavior in online shopping
  via reinforcement learning.
\newblock \emph{arXiv preprint arXiv:2507.17842}.

\end{thebibliography}
\appendix

\section{Ethics Statement}
\label{app:ethic_state}
The ethical statement about this work is as follows:
\begin{itemize}
    [itemsep=4pt,topsep=0pt,parsep=0pt,leftmargin=9pt]
    \item \textbf{Data Collection}: All product information is collected with authorization from the e-commerce platform and complies with local legal and platform policies. The data contains no personal privacy information.
    \item \textbf{Annotation Recruitment}: Since our study is conducted in Chinese, we recruited native Chinese speakers for annotation, each with at least a bachelor's degree. We set per-task payments based on task duration and difficulty. The payment is higher than the local minimum wage requirement. Annotators only need to work online, and the annotation process does not involve any harm to their physical or mental health.
\end{itemize}
\begin{table*}[!t]
    \centering
    \scriptsize
\renewcommand{\arraystretch}{1.2}
\setlength{\tabcolsep}{5mm}{
    \begin{tabular}{lll}
    \toprule
        \textbf{Domain} & \textbf{Full Name} & \textbf{Description} \\ 
        \midrule
        Home & Home \& Living & \makecell[l]{Furniture, décor, bedding, kitchenware, and home improvement materials.} \\ 
        Supplies & Industrial \& Farm Supplies & \makecell[l]{Agricultural goods, machinery, raw materials, tools, and industrial components.} \\ 
        Leisure & \makecell[l]{Leisure\& Entertainment\\ \& Education} & \makecell[l]{Books, toys, musical instruments, games, collectibles, and cultural products.} \\ 
        Sports & \makecell[l]{Sports \& Outdoor \& Travel} & \makecell[l]{Sporting goods, outdoor equipment, bicycles, vehicles, and automotive accessories.} \\ 
        Beauty & Beauty \& Personal Care \& Health & \makecell[l]{Cosmetics, personal hygiene, medical devices, supplements, and wellness products.} \\ 
        Appliances & Home Appliances \& Electronics & \makecell[l]{Appliances, computers, cameras, mobile phones, and electronic accessories.} \\ 
        Food & Food \& Drinks & \makecell[l]{Packaged foods, beverages, fresh produce, tea, coffee,  and liquor.} \\ 
        Clothing & Clothing \& Shoes \& Accessories & \makecell[l]{Apparel, footwear, fashion accessories, bags, jewelry, and watches.} \\ 
        Kids & Maternity \& Baby \& Kids & \makecell[l]{Baby products, kids’ clothing and shoes, maternity \\wear, and diapers.} \\ 
        Services & Local \& General Services & \makecell[l]{Local life services, travel, events, real estate, custom design, and public welfare.} \\ 
        Digital & Online Services \& Digital Goods & \makecell[l]{Game cards, telecom recharge, e-vouchers, software, and online service packages.} \\ 
        Secondhand & Second-hand \& Auctions & \makecell[l]{Pre-owned goods, used electronics, and judicial or live-auction items.} \\ 
        \bottomrule
    \end{tabular}
}
    \caption{The introduction to the 12 domains of ShopSimulator.}
\label{tab:domain_intro}
\end{table*}

\begin{table*}[!t]
\centering
\footnotesize
\setlength{\tabcolsep}{5mm}{
\begin{adjustbox}{width=\textwidth, keepaspectratio}
\begin{tabular}{lll}
\toprule
\textbf{} & \textbf{Error Type} & \textbf{Description} \\ \midrule

\multirow{4}{*}{\makecell[l]{Search\\Action}} 
 & Ignored key attribute & The agent ignored known key product attributes, reducing search accuracy. \\
 & Abandoned high-match result & The agent abandoned a highly matching search result and restarted searching. \\
 & Repeated similar query & The agent repeated a similar search query without changes in the context. \\
 & Others & Other unreasonable search-related behaviors. \\ \cmidrule{2-3}

\multirow{5}{*}{\makecell[l]{Click\\Action}}
 & Violated hard requirement & The agent clicked an item that violated a hard requirement from the shopper. \\
 & Unconfirmed key attribute & The agent selected a product specification without confirming a key attribute. \\
 & Nonexistent button & The agent clicked a button that does not exist on the current page. \\
 & Retried rejected attribute & The agent repeatedly tried an attribute that the shopper had already rejected. \\
 & Others & Other unreasonable click-related behaviors. \\ \cmidrule{2-3}

\multirow{3}{*}{\makecell[l]{Buy Now\\Action}}
 & No detail confirmation & The agent proceeded to purchase without confirming important product details. \\
 & Purchase after rejection & The agent completed a purchase even after the shopper rejected or changed the request. \\
 & Others & Other unreasonable buy-related behaviors. \\ \cmidrule{2-3}

\multirow{4}{*}{\makecell[l]{Ask Shopper\\Action}}
 & Do not ask when info missing & The agent did not ask for key information when it was missing, and proceeded with actions. \\
 & Over-confirmed known info & The agent repeatedly confirmed information that was already known. \\
 & Asked after farewell & The agent continued to ask questions after the shopper said farewell. \\
 & Others & Other unreasonable ask-related behaviors. \\ \midrule

\multirow{3}{*}{Personalization}
 & Ignore personal info & The agent ignored available personalization data during decision-making. \\
 & Overinterpret personal info & The agent drew overly specific conclusions from personalization data. \\
 & Mix short- and long-term priorities & The agent confused immediate needs with shopper’s long-term preferences. \\ \midrule

\multirow{3}{*}{Shopper}
 & Adding extra intent & The shopper introduced additional intents unrelated to the original goal. \\
 & Distorting target intent & The shopper distorted or misrepresented the original goal’s intent. \\
 & Silent on key goal & The shopper did not respond to questions about key goal attributes. \\
\bottomrule

\end{tabular}
\end{adjustbox}
}
\caption{The description of error types in LLM agents, and of error types in LLM-simulated shoppers.}
\label{tab:error_type_description}
\end{table*}

\section{Potential Risk}
\label{app:potential_risk}
We discuss the potential risks as follows:
\begin{itemize}
[itemsep=4pt,topsep=0pt,parsep=0pt,leftmargin=9pt]
\item Our product catalog is collected from a June 2025 snapshot of the e-commerce platform, so the item distribution may reflect time- or platform-specific structure. Therefore, the agent may perform unevenly under distribution shift.
\item We do not use real user profiles; instead, we rely on LLM-synthesized profiles, which may encode biased or incomplete preferences and thus yield uneven performance across user groups.
\item Our instruction data relies on human annotation and review; subjective variation and consistency drift may introduce minor label noise.
\end{itemize}

\section{Details of ShopSimulator}
\label{app:env_details}

\subsection{Details of LLM Prompt}
\label{subapp:prompt_details}

Figure \ref{fig:agent_system_prompt} shows the system prompt designed to guide the LLM in acting as a shopping assistant, specifying the rules and steps for collecting user requirements during the conversation, searching for and clicking on products, and completing the purchase after confirming that the needs are met.

Figure \ref{fig:user_system_prompt} shows the system prompt designed to guide the LLM in acting as a shopper, requiring it to gradually disclose all details of its purchase goal to the assistant through natural conversation, and to refuse the purchase if the information is incomplete.

\subsection{Example of ShopSimulator}
\label{subapp:example_details}
We provide different types of examples as follows:
\begin{itemize}
[itemsep=4pt,topsep=0pt,parsep=0pt,leftmargin=9pt]
    \item In Table \ref{tab:domain_intro}, we introduce the 12 domains in ShopSimulator.
    \item Figures \ref{fig:product_case}, \ref{fig:task_case}, and \ref{fig:profile_case} respectively show examples of products, tasks, and user profiles used for personalization in ShopSimulator.
    \item Figures \ref{fig:single_turn_case_part1}, \ref{fig:single_turn_case_part2}, and \ref{fig:single_turn_case_part3} present the trajectory of an LLM agent in a single‑turn scenario, while Figure \ref{fig:multi_turn_case} and Figure \ref{fig:multi_turn_personlization_case} respectively present example trajectories for the multi‑turn and multi‑turn personalization scenarios.
\end{itemize}

\subsection{Details of Reward}
\label{subapp:reward_details}

Section \ref{subsec:reward} introduces the overall formulation of the reward function. Here we briefly describe the meaning and computation of each component, following the implementation in WebShop~\citep{yao2022webshop}:
\begin{itemize}
[itemsep=4pt,topsep=0pt,parsep=0pt,leftmargin=9pt]
    \item $R_{\text{type}}$: Measures the consistency between the purchased product’s category and the target category, based on three criteria: (1) whether the initial search query is exactly identical; (2) whether the category paths share at least two common nodes; and (3) whether the overlap ratio of title keywords exceeds 0.2. If any condition is satisfied, the coefficient is set to 1.0; otherwise it is set to 0.5, and further reduced to 0.1 or 0.0 when the title similarity drops below 0.1 or equals 0, respectively.
    \item $R_{\text{attribute}}$: Quantifies the proportion of semantic attributes (e.g., material, functionality, style) of the purchased product that match those required by the target. Fuzzy string matching is applied to determine direct matches; if a target attribute is not found in the product’s attribute list, it is additionally checked for occurrence within the product’s title or detailed description. The rate is computed as the number of matched attributes divided by the total number of target attributes.
    \item $R_{\text{option}}$: Evaluates the proportion of configuration options (e.g., color, size, capacity) of the purchased product that match the target’s required options, using fuzzy matching. The rate is given by the number of matched options divided by the total number of target options.
    \item $R_{\text{price}}$: Assigned a value of 1 if the purchased product’s price does not exceed the target price limit, and 0 otherwise.
\end{itemize}

\section{Details of Experiments}
\label{app:exp_details}

\subsection{Implementation of Training}
\label{subapp:rl_implement}
For SFT, we collect 6K successful trajectories from GPT‑4.1 on the training set, and fine‑tune Qwen3‑8B with a batch size of 32 and a learning rate of \(1.0\times 10^{-5}\) for 4 epochs.
For RL, we use the ROLL framework~\citep{wang2025reinforcement} with the GRPO algorithm~\citep{shao2024deepseekmath}, omitting the KL loss term to encourage exploration.
The maximum context is set to 32K tokens, and the learning rate to \(1.0\times 10^{-6}\). We trained for a total of 200 steps, with each step performing 8 trajectory rollouts for 32 samples.



\begin{figure*}
\begin{tcolorbox}[
colback=white!10!white,
colframe=black!75!black,
left=0.5pt, right=0.5pt,
boxsep=1pt,
title=System Prompt for Shopping Agent,
]
\vspace{-0.3cm}
\begin{lstlisting}
#### Task Description ####  
You are an intelligent shopping assistant **Agent** who needs to help the **Shopper** achieve their purchase goal. Your core responsibilities are:  

1. **Information Collection**: Ask the Shopper questions to gather key details -- product type, budget range, attribute preferences, specification requirements, etc.  
2. **Environment Interaction**: Perform precise searches in ShopEnv (search[keyword]) and valid clicks (click[value]).  
3. **Process Control**: Ensure product specifications are selected before purchase (must click the "product specifications" button before clicking buy).  
4. **Decision Optimization**: Adjust strategy dynamically based on real-time observations -- prioritize asking questions when information is incomplete, and take actions when information is sufficient.  
5. **Final Confirmation Mechanism**: Before deciding to purchase, confirm with the Shopper that the selected product fully matches their needs in features and attributes; if the Shopper refuses to buy or provides new information, you must not execute the purchase.  

#### Response Format ####  
Always return responses in the following format:  
Thought: Explain how you decide the next step  
Action_type: ask_shopper | interact_with_env  
Action_content: Detailed content  

- If action_type is ask_shopper:  
   - content should be an open-ended question (e.g., "What's your budget range for the product?").  
- If action_type is interact_with_env:  
   content must follow one of these formats:  
   - search[keyword] (e.g., search[wireless noise-cancelling headphones])  
   - click[value] (e.g., click[< pre])  
   - Note: click[buy now] means directly purchasing the product, not going to the purchase page.  

#### Interaction Rules ####  
1. **Search Strategy**:  
   - Keywords must include core attributes (brand/model/specs).  
   - Prioritize using detailed information provided by the Shopper.  
   - Use search[keyword] only when search function is available.  
2. **Click Strategy**:  
   - Values in click[value] must come from currently available clickable buttons (not historical ones).  
   - **Note**: click[buy now] means purchasing immediately, not navigating to the buy page.  
3. **Decision Logic**:  
   - Missing info -> ask_shopper  
   - Action possible -> interact_with_env  
   - Only one action per turn.  
   - If at the end of the conversation the user still has no satisfactory product, select and buy the item you consider most suitable -- no need to confirm with them.  
4. **Ending Rules**:  
   - Before ending the conversation / reaching the dialogue limit, you must interact with the environment to purchase the most suitable product for the user.  
   - If the Shopper says goodbye, do not speak to them again -- immediately buy the most suitable product in the current catalog without notifying or confirming with them.  

#### Notes ####  
1. Always use the required response format.  
2. Do not say goodbye to the user more than once.  
3. If you do not execute click[buy now] before reaching the turn limit, your task completion rate will be 0.  
4. If the environment does not change, check whether your action target exists in the current clickable buttons list. 
\end{lstlisting}
\vspace{-0.3cm}
\end{tcolorbox}
\vspace{-0.3cm}
\caption{The system prompt for LLM assistant (Chinese original, presented in English for universal readability).}
\label{fig:agent_system_prompt}
\end{figure*}

\begin{figure*}
\begin{tcolorbox}[
colback=white!10!white,
colframe=black!75!black,
left=0.5pt, right=0.5pt,
boxsep=1pt,
title=System Prompt for Acting as a User,
]
\vspace{-0.3cm}
\begin{lstlisting}
You are a simulated shopper trying to complete a product purchase task through a conversation with a customer service agent. Your task is:

1. You have a **specific purchase goal** (only you know it), but you won't reveal it at the start. Begin with a vague purchase intention and engage in a natural, multi-turn conversation with the agent.  
2. In the conversation, your main role is to answer the agent's questions to help them gradually understand your needs, until they can find and recommend the exact product you want.  
3. **Do not voluntarily provide detailed information -- wait for the agent to ask.**  
4. Before the conversation ends (i.e., before the agent proceeds to purchase), make sure the agent has obtained all features and attributes of your specific purchase goal -- no details can be missing.  
5. If some information has not yet been told to the agent, refuse the purchase and give a simple reason, then wait for further questions.  
6. Keep your language natural and conversational, as a real shopper would.  

Your purchase goal this round is: {goal}
\end{lstlisting}
\vspace{-0.3cm}
\end{tcolorbox}
\vspace{-0.3cm}
\caption{The system prompt for LLM-simulated shopper (Chinese original, presented in English for universal readability).}
\label{fig:user_system_prompt}
\end{figure*}

\begin{figure*}
\begin{tcolorbox}[
colback=white!10!white,
colframe=black!75!black,
left=0.5pt, right=0.5pt,
boxsep=1pt,
title=Product Example,
]
\vspace{-0.3cm}
\begin{lstlisting}
{
  "title": "Authentic YONEX YY badminton shoes, men's cushioning and wear-resistant badminton-specific wide-last sports shoes, women's version",
  "shop_name": "Miaojiang Sports & Outdoor Specialty Store",
  "domain": "Clothing, Shoes, Accessories",
  "first_category": "Athletic Shoes",
  "fine_category": "Badminton Shoes",
  "options": {
    "Color Options": ["SHB510WCR Black/Red (Wide last)", "SHB610WCR White/Navy (Wide last)", "SHB510WCR White/Blue (Wide last)", "SHB510WCR White (Wide last)", "SHB510WCR Silver/Gray (Wide last)"],
    "Size": ["43", "42", "44", "36", "38", "39", "37", "41", "40", "45"]
  },
  "pricing": 528.0,
  "attribute": ["Cushioning", "Wear-resistant", "Authentic", "Unisex"]
}
\end{lstlisting}
\vspace{-0.3cm}
\end{tcolorbox}
\vspace{-0.3cm}
\caption{An example for a product in ShopSimulator (Chinese original, presented in English for universal readability.)}
\label{fig:product_case}
\end{figure*}

\begin{figure*}
\begin{tcolorbox}[
colback=white!10!white,
colframe=black!75!black,
left=0.5pt, right=0.5pt,
boxsep=1pt,
title=Task Example,
]
\vspace{-0.3cm}
\begin{lstlisting}
{
    "instruction": "A friend said wide-last badminton shoes are more suitable for sports. Could you find me a genuine pair with cushioning and wear resistance? I prefer blue-and-white colors, which feel easy to match. I want a unisex style. The required size is EU 40, and the budget is within 550 yuan.",
    "target_product": "Authentic YONEX YY badminton shoes, men's cushioning and wear-resistant badminton-specific wide-last sports shoes, women's version",
    "target_options": {
        "Color Options":"SHB610WCR White/Navy (Wide last)",
        "Size": "40"
    },
    "target_attributes": ["Cushioning", "Wear-resistant", "Authentic", "Unisex"]
}
\end{lstlisting}
\vspace{-0.3cm}
\end{tcolorbox}
\vspace{-0.3cm}
\caption{An example for a task in ShopSimulator.(Chinese original, presented in English for universal readability.)}
\label{fig:task_case}
\end{figure*}

\begin{figure*}
\begin{tcolorbox}[
colback=white!10!white,
colframe=black!75!black,
left=0.5pt, right=0.5pt,
boxsep=1pt,
title=Personal Profile Example,
]
\vspace{-0.3cm}
\begin{lstlisting}
{
  "Transaction Characteristics": {"Coupon Usage Rate": 0.32, "Repeat Purchase Rate": 0.68, "Average Order Value": 177.56, "Preferred Payment Method": "Alipay", "Is Promotion-Sensitive": false, "Spending in Last 30 Days": 1420.5, "Orders in Last 90 Days": 8},
  "Demographics": {"Membership Level": "Gold Member","Age Range": "25-34","Gender": "Female","Spending Level": "Medium"},
  "Interests and Preferences": {
    "Brand Preferences": [
      { "Preference Level": "High", "Brand Name": "YONEX" },
      { "Preference Level": "High", "Brand Name": "ASICS" },
      { "Preference Level": "Medium", "Brand Name": "Li-Ning" },
      { "Preference Level": "Low", "Brand Name": "Wilson" },
      { "Preference Level": "Medium", "Brand Name": "Lululemon" }],
    "Product Attribute Preferences": {
      "Price Range": {"Max": 800, "Min": 200},
      "Features": ["Cushioning", "Wear-resistant", "Stable Support"],
      "Size Preferences": {"Tops": "", "Underwear": "", "Dresses/Skirts": "", "Pants": "", "Shoe Size": "40"},
      "Materials": ["Breathable Mesh", "Synthetic Leather"],
      "Colors": ["Blue", "White", "Gray"],
      "Styles": ["Minimalist Sport", "Techwear"]},
    "Category Preferences": {"Books & Media": "Low", "Pet Supplies": "Medium", "Smart Home": "Medium", "Apparel, Shoes & Bags": "Medium", "Mother & Baby": "None", "Beauty & Skincare": "Low", "Sports & Outdoors": "High"}},
  "Location Information": {"District": "Nanshan District","City": "Shenzhen","Time Zone": "Asia/Shanghai","Province": "Guangdong"},
  "Last Updated": "2024-05-25T09:41:12Z",
  "Registration Time": "2021-03-18T14:22:36Z",
  "Registration Channel": "App",
  "User ID": "U98421537",
  "User Tags": ["Weekend badminton club organizer","Mid-to-high-frequency sports spending","Brand and function oriented","Blue-and-white color preference","Low promotion sensitivity","Gold member"],
  "Behavioral Features": {"Commonly Used Device": "iOS smartphone", "Average Daily Browsing Duration": 3900, "Shops Followed in Last 14 Days": [], "Items Added to Cart in Last 14 Days": [], "Search Keywords in Last 14 Days": ["Cushioned wear-resistant badminton-specific shoes", "Authentic YONEX wide-last sports shoes"], "Items Favorited in Last 14 Days": [], "Active Hours": [7, 12, 22], "Visits in Last 7 Days": 18}
}
\end{lstlisting}
\vspace{-0.3cm}
\end{tcolorbox}
\vspace{-0.3cm}
\caption{An example for a user profile for personalization.(Chinese original, presented in English for universal readability.)}
\label{fig:profile_case}
\end{figure*}

\begin{figure*}
\begin{tcolorbox}[
colback=white!10!white,
colframe=black!75!black,
left=0.5pt, right=0.5pt,
boxsep=1pt,
title=Single Turn Case (Part 1),
]
\vspace{-0.3cm}
\begin{lstlisting}
# User
A friend said wide-last badminton shoes are more suitable for sports. Could you find me a genuine pair with cushioning and wear resistance? I prefer blue-and-white colors, which feel easy to match. I want a unisex style. The required size is EU 40, and the budget is within 550 yuan.

# Environment Initial Observation
WebShop [SEP] Instruction: [SEP] A friend said wide-last badminton shoes are more suitable for sports. Could you find me a genuine pair with cushioning and wear resistance? I prefer blue-and-white colors, which feel easy to match. I want a unisex style. The required size is EU 40, and the budget is within 550 yuan. [SEP] Search
Is search available: True
Clickable buttons: []

# Agent
Thought: The current page has no product list, so I can only find suitable products via search. Based on the user's needs, the search keywords should cover "wide last," "badminton shoes," "blue and white," "cushioning and wear-resistant," and "genuine" to maximize the chance of finding the most relevant items.
Action: search[wide last blue white badminton shoes cushioning wear-resistant genuine]

# Env
Instruction: [SEP] A friend said wide-last badminton shoes are more suitable for sports. Could you find me a genuine pair with cushioning and wear resistance? I prefer blue-and-white colors, which feel easy to match. I want a unisex style. The required size is EU 40, and the budget is within 550 yuan. [SEP] Back to Search [SEP] Page 1 (Total results: 150) [SEP] Next > [SEP] 724988974873 [SEP] Authentic YONEX YY badminton shoes, men's cushioning and wear-resistant, badminton-specific, wide last sports shoes, women's version [SEP] 528.0 to 660.0 [SEP] 674239829358 [SEP] Official authentic VICTOR 9200TD professional badminton shoes, wide last, cushioning, anti-slip, anti-torsion, unisex [SEP] 518.0 [SEP] 775702903455 [SEP] Li-Ning badminton shoes, authentic All-round King, unisex professional training, 2024 new, cushioning, wide last, sports/tennis shoes [SEP] 310.0 [SEP] 899453734819 [SEP] VICTOR badminton shoes P8500CLS, cushioning, anti-torsion, unisex, breathable, wide last, anti-slip [SEP] 398.0 [SEP] 925395439375 [SEP] VICTOR badminton shoes A170 II, ultra-light cushioning, unisex, wide last [SEP] 213.0 [SEP] 678221250065 [SEP] kawasaki/Kawasaki anti-slip wear-resistant badminton shoes, unisex training shoes, cushioning, breathable, lightweight, authentic [SEP] 223.0 [SEP] 827282336955 [SEP] 2024 new Li-Ning authentic badminton shoes, unisex, Zhanji 2.0, wear-resistant, cushioning, competition/training [SEP] 899.0 [SEP] 894201472400 [SEP] Brand clearance P yy badminton shoes, unisex, anti-slip, wear-resistant, tennis training shoes, cushioning, breathable, professional sports shoes [SEP] 130.0 [SEP] 771384727942 [SEP] Warrior badminton shoes, unisex breathable sports shoes, anti-slip, wear-resistant, cushioning; table tennis/volleyball training shoes [SEP] 350.0 to 369.0 [SEP] 726975937654 [SEP] Li-Ning Zhanji III LITE women's cushioning wear-resistant support badminton training shoes AYTR026-2 [SEP] 229.0 to 299.0 [SEP] 817948252002 [SEP] Venson badminton shoes VS169, wear-resistant, anti-slip, cushioning, high rebound, wrapping, breathable sports shoes [SEP] 199.0 [SEP] 824554389645 [SEP] Li-Ning badminton shoes, unisex, anti-slip, cushioning, breathable, wear-resistant, casual/professional competition training tennis shoes [SEP] 530.0 [SEP] 850124091149 [SEP] kawasaki/Kawasaki badminton shoes, unisex professional cushioning wear-resistant breathable sports training shoes "Puff" [SEP] 349.0 [SEP] 886630046021 [SEP] Li-Ning badminton shoes Thunder beng tech pink, unisex, cushioning, wear-resistant, rebound, competition shoes AYAS018 [SEP] 469.0 to 899.0 [SEP] 907578716897 [SEP] Li-Ning badminton shoes Ground Flight LITE, unisex, cushioning, anti-slip, wear-resistant, training sports shoes AYTT003 [SEP] 499.0 [SEP] 895792505809 [SEP] Kawasaki badminton shoes, unisex, wear-resistant, breathable, anti-slip, cushioning sports shoes, Flying youth model, flagship store [SEP] 500.0 [SEP] 816090104446 [SEP] Li-Ning Thunder NC badminton shoes, unisex, anti-torsion protection, wear-resistant, cushioning, professional competition shoes AYZU015 [SEP] 804.0 [SEP] 761658225777 [SEP] Li-Ning badminton shoes Ground Flight 2 SE, unisex, comfortable cushioning rebound, wear-resistant, professional competition shoes [SEP] 899.0 [SEP] 867510242711 [SEP] ASICS badminton shoes, men's UPCOURT Cross 6 professional sports shoes, shock-absorbing, anti-slip, wear-resistant, authentic [SEP] 590.0 [SEP] 724352099396 [SEP] YONEX tennis shoes F5, wear-resistant, cushioning, breathable, F5 unisex professional tennis sports shoes [SEP] 698.0
Is search available: False
Clickable buttons: ["back to search", "next >", "724988974873", "674239829358", "775702903455", "899453734819", "925395439375", "678221250065", "827282336955", "894201472400", "771384727942", "726975937654", "817948252002", "824554389645", "850124091149", "886630046021", "907578716897", "895792505809", "816090104446", "761658225777", "867510242711", "724352099396"]
\end{lstlisting}
\vspace{-0.3cm}
\end{tcolorbox}
\caption{A trajectory (part 1/3) of an agent interacting with the ShopSimulator environment in a single‑turn scenario. (Chinese original, presented in English for universal readability.)}
\label{fig:single_turn_case_part1}
\end{figure*}

\begin{figure*}
\begin{tcolorbox}[
colback=white!10!white,
colframe=black!75!black,
left=0.5pt, right=0.5pt,
boxsep=1pt,
title=Single Turn Case (Part 2),
]
\vspace{-0.3cm}
\begin{lstlisting}
# Agent
Thought: The search results show many badminton shoes, with multiple items containing keywords like "wide last," "cushioning," "wear-resistant," and "authentic." I need to further filter for "blue and white," unisex, under 550 yuan, and with size 40 available. Priorities are items whose appearance and description best match. Comparing, YONEX, VICTOR, and Li-Ning have wide-last and unisex options; Li-Ning All-round King (310 yuan), VICTOR P8500CLS (398 yuan), and VICTOR 9200TD (518 yuan) are within budget. The YONEX YY (528 yuan) mentions "wide" and women's option; I should verify whether it has blue-and-white and size 40. Next, click the YONEX YY badminton shoes to enter details and verify color and size.
Action: click[724988974873]

# Env
Instruction: [SEP] A friend said wide-last badminton shoes are more suitable for sports. Could you find me a genuine pair with cushioning and wear resistance? I prefer blue-and-white colors, which feel easy to match. I want a unisex style. The required size is EU 40, and the budget is within 550 yuan. [SEP] Back to Search [SEP] < Prev [SEP] Color options [SEP] shb510wcr Black/Red (Wide last) [SEP] shb610wcr White/Navy (Wide last) [SEP] shb510wcr White/Blue (Wide last) [SEP] shb510wcr White (Wide last) [SEP] shb510wcr Silver/Gray (Wide last) [SEP] Size [SEP] 43 [SEP] 42 [SEP] 44 [SEP] 36 [SEP] 38 [SEP] 39 [SEP] 37 [SEP] 41 [SEP] 40 [SEP] 45 [SEP] Authentic YONEX YY badminton shoes, men's cushioning wear-resistant badminton-specific wide last sports shoes, women's version [SEP] Price: 528.0 to 660.0 [SEP] Store: Miaojiang Sports & Outdoor Specialty Store [SEP] Description [SEP] Features [SEP] Reviews [SEP] Buy Now
Is search available: False
Clickable buttons: ["back to search", "< prev", "description", "features", "reviews", "buy now", "shb510wcr Black/Red (Wide last)", "shb610wcr White/Navy (Wide last)", "shb510wcr White/Blue (Wide last)", "shb510wcr White (Wide last)", "shb510wcr Silver/Gray (Wide last)", "43", "42", "44", "36", "38", "39", "37", "41", "40", "45"]

# Agent
Thought: This page offers multiple color options and sizes, all wide last. The user specified blue-and-white and size 40, and the 528 yuan price fits the budget. The best match is "shb510wcr White/Blue (Wide last)." I should select that color, then choose size 40 to prepare for purchase.
Action: click[shb510wcr White/Blue (Wide last)]

# Env
Instruction: [SEP] A friend said wide-last badminton shoes are more suitable for sports. Could you find me a genuine pair with cushioning and wear resistance? I prefer blue-and-white colors, which feel easy to match. I want a unisex style. The required size is EU 40, and the budget is within 550 yuan. [SEP] Back to Search [SEP] < Prev [SEP] Color options [SEP] shb510wcr Black/Red (Wide last) [SEP] shb610wcr White/Navy (Wide last) [SEP] shb510wcr White/Blue (Wide last) [SEP] shb510wcr White (Wide last) [SEP] shb510wcr Silver/Gray (Wide last) [SEP] Size [SEP] 43 [SEP] 42 [SEP] 44 [SEP] 36 [SEP] 38 [SEP] 39 [SEP] 37 [SEP] 41 [SEP] 40 [SEP] 45 [SEP] Authentic YONEX YY badminton shoes, men's cushioning wear-resistant badminton-specific wide last sports shoes, women's version [SEP] Price: 528 [SEP] Store: Miaojiang Sports & Outdoor Specialty Store [SEP] Description [SEP] Features [SEP] Reviews [SEP] Buy Now
Is search available: False
Clickable buttons: ["back to search", "< prev", "description", "features", "reviews", "buy now", "shb510wcr Black/Red (Wide last)", "shb610wcr White/Navy (Wide last)", "shb510wcr White/Blue (Wide last)", "shb510wcr White (Wide last)", "shb510wcr Silver/Gray (Wide last)", "43", "42", "44", "36", "38", "39", "37", "41", "40", "45"]

# Agent
Thought: "White/Blue (Wide last)" color and size 40 are selected. The price is 528 yuan, within budget, and the description indicates authentic, wide last, cushioning, wear-resistant, unisex--meeting all requirements. Now proceed to purchase.
Action: click[40]
\end{lstlisting}
\vspace{-0.3cm}
\end{tcolorbox}
\caption{A trajectory (part 2/3) of an agent interacting with the ShopSimulator environment in a single‑turn scenario. (Chinese original, presented in English for universal readability.)}
\label{fig:single_turn_case_part2}
\end{figure*}

\begin{figure*}
\begin{tcolorbox}[
colback=white!10!white,
colframe=black!75!black,
left=0.5pt, right=0.5pt,
boxsep=1pt,
title=Single Turn Case (Part 3),
]
\vspace{-0.3cm}
\begin{lstlisting}
# Env
Instruction: [SEP] A friend said wide-last badminton shoes are more suitable for sports. Could you find me a genuine pair with cushioning and wear resistance? I prefer blue-and-white colors, which feel easy to match. I want a unisex style. The required size is EU 40, and the budget is within 550 yuan. [SEP] Back to Search [SEP] < Prev [SEP] Color options [SEP] shb510wcr Black/Red (Wide last) [SEP] shb610wcr White/Navy (Wide last) [SEP] shb510wcr White/Blue (Wide last) [SEP] shb510wcr White (Wide last) [SEP] shb510wcr Silver/Gray (Wide last) [SEP] Size [SEP] 43 [SEP] 42 [SEP] 44 [SEP] 36 [SEP] 38 [SEP] 39 [SEP] 37 [SEP] 41 [SEP] 40 [SEP] 45 [SEP] Authentic YONEX YY badminton shoes, men's cushioning wear-resistant badminton-specific wide last sports shoes, women's version [SEP] Price: 528 [SEP] Store: Miaojiang Sports & Outdoor Specialty Store [SEP] Description [SEP] Features [SEP] Reviews [SEP] Buy Now

Is search available: False

Clickable buttons: ["back to search", "< prev", "description", "features", "reviews", "buy now", "shb510wcr Black/Red (Wide last)", "shb610wcr White/Navy (Wide last)", "shb510wcr White/Blue (Wide last)", "shb510wcr White (Wide last)", "shb510wcr Silver/Gray (Wide last)", "43", "42", "44", "36", "38", "39", "37", "41", "40", "45"]

# Agent
Thought: The "White/Blue (Wide last)" color and size 40 are selected, the price is 528 yuan within budget, and the product is authentic, wide last, cushioned, wear-resistant, and unisex--fully meeting the user's needs. Proceed to purchase.
Action: click[buy now]
\end{lstlisting}
\vspace{-0.3cm}
\end{tcolorbox}
\caption{A trajectory (part 3/3) of an agent interacting with the ShopSimulator environment in a single‑turn scenario. (Chinese original, presented in English for universal readability.)}
\label{fig:single_turn_case_part3}
\end{figure*}

\begin{figure*}
\begin{tcolorbox}[
colback=white!10!white,
colframe=black!75!black,
left=0.5pt, right=0.5pt,
boxsep=1pt,
title=Multi-Turn Case,
]
\vspace{-0.3cm}
\begin{lstlisting}
[User] I want to buy a pair of badminton shoes, preferably cushioned and wear-resistant, and they must be authentic.

[Agent] Hello! I'll help you choose badminton shoes. To recommend the most suitable product, what's your budget range? Also, do you have any preferred brands (e.g., Li-Ning, Victor, YONEX)? And what is your shoe size?

[User] Budget within 550 yuan, no specific brand preference, but they must be authentic. Size is EU 40.

(Omitting a series of Agent-shopping environment interaction details, where the Agent searches based on the user's budget, size, and functional needs, checks multiple brands (YONEX, Li-Ning, Kawasaki), filters out items that don't meet budget or color requirements, and finally locks in the Li-Ning Ground Flight LITE at 499 yuan, size 40, black-and-white colorway.)

[Agent] I found a great option for you: Li-Ning Ground Flight LITE badminton shoes, size 40, black with standard white, priced at 499 yuan. They feature cushioning, anti-slip, and wear resistance, and belong to Li-Ning's professional badminton line, within your budget. How does this sound?

[User] This colorway is black and white, but I prefer blue-and-white. Could you find that colorway? Also, I've heard wide-last versions are more exercise-friendly--do you have a wide-last option?

(Omitting a series of Agent-shopping environment interaction details, where the Agent re-searches for the new "blue-and-white + wide last" requirements, browses YONEX, VICTOR, and Li-Ning wide-last models, and ultimately finds the VICTOR P8500CLS at 398 yuan, wide last, Glacier Blue colorway, size 40.)

[Agent] I found a pair that matches your needs well: VICTOR P8500CLS, wide-last version, Glacier Blue colorway, size 250 (equivalent to EU 40), priced at 398 yuan. It offers cushioning, anti-torsion, breathability, and anti-slip, a professional wide-last badminton shoe within your budget. What do you think?

[User] This pair is wide last, and the color is close to the blue-and-white I like, but I want a blue-and-white colorway, preferably a white-and-blue combo. Also, can you confirm whether it's a unisex model?

(Omitting a series of Agent-shopping environment interaction details, where the Agent continues searching for a blue-and-white, wide-last, unisex badminton shoe, filters multiple items, and finds the YONEX SHB520WCR: wide version, white-and-blue colorway, size 40, unisex, 400 yuan.)

[Agent] Great! I found a pair that fully meets your requirements: YONEX SHB520WCR, wide version, white-and-blue colorway (exactly what you want), size 40, unisex, priced at 400 yuan. It features cushioning and wear resistance and is a new model from YONEX, well within your budget. How about this one?

[User] The colorway, wide-last design, and price meet my expectations, but the exact model I want is SHB510WCR White/Blue (Wide last). Can you confirm size 40 is available? Also, please confirm there's an authenticity guarantee.

(Omitting a series of Agent-shopping environment interaction details, where the Agent precisely searches the specified model, enters an authorized YONEX store, finds SHB510WCR White/Blue (Wide last) in size 40 at 528 yuan, and confirms authenticity guarantee, meeting all requirements.)

[Agent] Found it! I located your exact specified model: YONEX SHB510WCR White/Blue (Wide last), size 40, priced at 528 yuan. It's from Miaojiang Sports & Outdoor Specialty Store with an authenticity guarantee. The 528 yuan price is within your 550 yuan budget, with cushioning and wear resistance--exactly the wide-last design and blue-and-white colorway you requested. Would you like to proceed with this purchase?

[User] Yes, this YONEX SHB510WCR White/Blue (Wide last) fully meets my needs: - Correct model (SHB510WCR) - Blue-and-white colorway - Wide-last design - Size is EU 40 - Unisex - Features cushioning and wear resistance - Authenticity guarantee - Price 528 yuan, within the 550 yuan budget

All conditions are met. I confirm the purchase now. Thank you for your patient help!

[Agent] click[buy now]
\end{lstlisting}
\vspace{-0.3cm}
\end{tcolorbox}
\caption{A trajectory of an agent interacting with the user and environment in a multi‑turn scenario. (Chinese original, presented in English for universal readability.)}
\label{fig:multi_turn_case}
\end{figure*}

\begin{figure*}
\begin{tcolorbox}[
colback=white!10!white,
colframe=black!75!black,
left=0.5pt, right=0.5pt,
boxsep=1pt,
title=Multi-Turn \& Personalization Case,
]
\vspace{-0.3cm}
\begin{lstlisting}
# Initial Observation
You are an intelligent shopping assistant Agent. I will provide you with the user's personal profile (including preferences, budget, usage scenarios, and purchase history).
Your goal is to, through multi-turn interactions with the shopping environment and the user, ultimately help the user purchase the product that best meets their needs.

Please pay special attention:
- For information that is vague, missing, or uncertain, ask the user follow-up questions to confirm and complete it.
- The product catalog contains many similar items, so you need to conduct comprehensive analysis and make precise judgments based on the personal profile plus real-time environment observations.

The user's personal profile is:

{
  "Transaction Characteristics": {"Coupon Usage Rate": 0.32, "Repeat Purchase Rate": 0.68, "Average Order Value": 177.56, "Preferred Payment Method": "Alipay", "Is Promotion-Sensitive": false, "Spending in Last 30 Days": 1420.5, "Orders in Last 90 Days": 8},
  "Demographics": {"Membership Level": "Gold Member","Age Range": "25-34","Gender": "Female","Spending Level": "Medium"},
  "Interests and Preferences": {
    "Brand Preferences": [
      { "Preference Level": "High", "Brand Name": "YONEX" },
      { "Preference Level": "High", "Brand Name": "ASICS" },
      { "Preference Level": "Medium", "Brand Name": "Li-Ning" },
      { "Preference Level": "Low", "Brand Name": "Wilson" },
      { "Preference Level": "Medium", "Brand Name": "Lululemon" }],
    "Product Attribute Preferences": {
      "Price Range": {"Max": 800, "Min": 200},
      "Features": ["Cushioning", "Wear-resistant", "Stable Support"],
      "Size Preferences": {"Tops": "", "Underwear": "", "Dresses/Skirts": "", "Pants": "", "Shoe Size": "40"},
      "Materials": ["Breathable Mesh", "Synthetic Leather"],
      "Colors": ["Blue", "White", "Gray"],
      "Styles": ["Minimalist Sport", "Techwear"]},
    "Category Preferences": {"Books & Media": "Low", "Pet Supplies": "Medium", "Smart Home": "Medium", "Apparel, Shoes & Bags": "Medium", "Mother & Baby": "None", "Beauty & Skincare": "Low", "Sports & Outdoors": "High"}},
  "Location Information": {"District": "Nanshan District","City": "Shenzhen","Time Zone": "Asia/Shanghai","Province": "Guangdong"},
  "Last Updated": "2024-05-25T09:41:12Z",
  "Registration Time": "2021-03-18T14:22:36Z",
  "Registration Channel": "App",
  "User ID": "U98421537",
  "User Tags": ["Weekend badminton club organizer","Mid-to-high-frequency sports spending","Brand and function oriented","Blue-and-white color preference","Low promotion sensitivity","Gold member"],
  "Behavioral Features": {"Commonly Used Device": "iOS smartphone", "Average Daily Browsing Duration": 3900, "Shops Followed in Last 14 Days": [], "Items Added to Cart in Last 14 Days": [], "Search Keywords in Last 14 Days": ["Cushioned wear-resistant badminton-specific shoes", "Authentic YONEX wide-last sports shoes"], "Items Favorited in Last 14 Days": [], "Active Hours": [7, 12, 22], "Visits in Last 7 Days": 18}
}

[User]
I want to buy a pair of badminton shoes, preferably in a white-and-blue colorway for a clean look.

[Agent] 
Thought: The user clearly wants badminton shoes and prefers a white-and-blue colorway. From the personal profile, I see a high preference for YONEX and ASICS, shoe size 40, a price range of 200-800 yuan, and feature preferences including cushioning, wear resistance, and stable support. The user's search keywords in the last 14 days also include "cushioned wear-resistant badminton-specific shoes" and "authentic YONEX wide-last sports shoes," further confirming the need. The search function is available now, so I should directly search for badminton shoes that match the user's requirements.
Action:search[YONEX badminton shoes white-and-blue colorway cushioning wear-resistant size 40]

......
\end{lstlisting}
\vspace{-0.3cm}
\end{tcolorbox}
\caption{A trajectory of an agent interacting with the user and environment in the Multi-Turn \& Personalization scenario. (Chinese
original, presented in English for universal readability.)}
\label{fig:multi_turn_personlization_case}
\end{figure*}

\end{document}